\documentclass[10pt,journal]{IEEEtran}

\newlength{\tocsep} 
\usepackage{color}
\usepackage{pdfpages}
\usepackage{multirow}
\usepackage{amsfonts, amssymb, amsmath}
\usepackage{hyperref}
\usepackage{subfigure}
\hypersetup{colorlinks = true, linkcolor = blue, filecolor = magenta, urlcolor = cyan,}
\usepackage{lettrine}
\usepackage{lipsum} 
\usepackage{setspace}
\usepackage{graphicx}
\usepackage{booktabs}
\usepackage{comment}
\usepackage[linesnumbered,ruled]{algorithm2e}
\usepackage{listings}
\usepackage{siunitx}
\usepackage{gensymb}
\usepackage[dvipsnames]{xcolor}
\usepackage[noadjust]{cite}
\graphicspath{ {figures/} }
\DeclareMathOperator*{\argmax}{argmax}
\DeclareMathOperator*{\argmin}{argmin}

\graphicspath{ {figures/} }

\newcommand{\wn}{\bar{\nu}}
\newcommand{\Hmat}{\mathbf{H}}
\newcommand{\Mmat}{\mathbf{M}}
\newcommand{\Fmat}{\mathbf{F}}
\newcommand{\Smat}{\mathbf{S}}
\newcommand{\Lmat}{\mathbf{L}}
\newcommand{\Bmat}{\mathbf{B}}
\newcommand{\Qmat}{\mathbf{Q}}
\newcommand{\Rmat}{\mathbf{R}}

\begin{document}
\title{Hyperspectral-Multispectral Image Fusion with Weighted LASSO}


\author{Nguyen~Tran,~\IEEEmembership{Student Member, IEEE}, Rupali Mankar,~\IEEEmembership{Student Member, IEEE},
        David~Mayerich,~\IEEEmembership{Senior Member, IEEE}, Zhu~Han, \IEEEmembership{Fellow, IEEE}
}
\markboth{IEEE Journal}%
{Shell \MakeLowercase{\textit{et al.}}: Bayesian Inference Regularization for Hyperspectral and Multispectral Image Fusion}
\maketitle
\begin{abstract}
Spectral imaging enables spatially-resolved identification of materials in remote sensing, biomedicine, and astronomy. However, acquisition times require balancing spectral and spatial resolution with signal-to-noise. Hyperspectral imaging provides superior material specificity, while multispectral images are faster to collect at greater fidelity. We propose an approach for fusing hyperspectral and multispectral images to provide high-quality hyperspectral output. The proposed optimization leverages the least absolute shrinkage and selection operator (LASSO) to perform variable selection and regularization. Computational time is reduced by applying the alternating direction method of multipliers (ADMM), as well as initializing the fusion image by estimating it using maximum a posteriori (MAP) based on Hardie's method. We demonstrate that the proposed sparse fusion and reconstruction provides quantitatively superior results when compared to existing methods on publicly available images. Finally, we show how the proposed method can be practically applied in biomedical infrared spectroscopic microscopy.
\end{abstract}


\flushbottom 

\thispagestyle{empty} 



\addcontentsline{toc}{section}{\hspace*{-\tocsep}Introduction} 

\newcommand\tab[1][1cm]{\hspace*{#1}}
\newcommand{\norm}[1]{\left\lVert#1\right\rVert}
\section{Introduction}
Hyperspectral imaging (HSI) combines spectroscopy with traditional digital imaging to acquire a localized material spectrum at each pixel. Spectral images are acquired using either hyperspectral techniques \cite{dale2013hyperspectral, manolakis2016hyperspectral, pahlow2018application}, which provide high spectral resolution, or multi-spectral techniques \cite{zhang2014visible}, which provide a limited subset of spectral bands. Hyperspectral images provide the best molecular specificity, however their high acquisition time imposes limitations in spatial resolution and image quality, making adjacent image features difficult to distinguish even though they may be chemically distinct. Multispectral imaging (MSI) employs tunable sources or filters to provide a subset of spectral samples in a much shorter time. Multi-spectral images generally provide greater spatial resolution at the expense of chemical specificity, particularly when the compounds within the image are unknown \textit{a priori}. We propose an approach that relies on fusing multi-modal images acquired using both HSI and MSI techniques to generate images with high spatial resolution, molecular specificity, and signal-to-noise (SNR).

Infrared (IR) microscopy is used to identify the distribution of chemical constituents in samples, such as cancer biopsies \cite{amenabar2017hyperspectral,stummer20135,petibois2010clinical,panasyuk2012medical}, without the need for destructive labeling \cite{nazeer2017infrared,kollermann2008expression, 4541041}. However, Fourier transform infrared (FTIR) spectroscopic imaging \cite{movasaghi2008fourier,fan2012fourier} is impractical for applications that require large samples (eg. \SI{2}{\centi\meter^2}) or high spatial resolution ($\approx$\SI{1}{\micro\meter}) \cite{lu2014medical} due to limits in light source intensity, mercury cadmium telluride (MCT) detector resolution, and diffraction-limited optics. While the recent availability of tunable quantum cascade lasers (QCLs) allows discrete-frequency infrared (DFIR) \cite{hughes2016introducing, tiwari2016towards}, this comes at the cost of spectral resolution, since individual bands must be selected for imaging \textit{a priori}. In addition, coherent illumination introduces imaging artifacts such as fringing.

Remote sensing applications leverage visible and near-infrared (NIR) imaging that have similar trade-offs. The Hyperion Imaging Spectrometer on an EO-01 satellite covers a \SIrange{400}{2500}{\nano\meter} spectral range with a spatial resolution of \SI{30}{\meter} \cite{EO01}. The GeoEye-1 satellite sensor provides a much higher spatial resolution (\SI{1.84}{\meter}) with a limited spectral range of \SIrange{450}{920}{\nano\meter} \cite{GEOEYE1}.

Several methods have been proposed for multi-modal image fusion \cite{ghassemian2016review} including pansharpening, dictionary learning, and convex optimization. Pansharpening fuses an multispectral (MS) tensor with a single high-resolution panchromatic image (PAN) \cite{huang2015new, vicinanza2015pansharpening, zhang2009noise}. This is typically easier than fusing two hyperspectral images, which is computationally prohibitive with existing pansharpening algorithms. Dictionary learning is based on linear spectral unmixing \cite{wei2015hyperspectral, chen2011hyperspectral} with the following assumptions: (1) multiple scattering between compounds is insignificant, (2) spectral components are discrete and readily separable, and (3) each constituent spectrum is known \cite{bioucas2012hyperspectral}. Dictionary learning relies on selecting a support tensor containing sufficient sparsity and density to remove noise while accounting for all molecular constituents. The support tensor is learned through Bayesian inference \cite{wei2015hyperspectral}, which provides a good prediction if the MSI and HSI have signal-to-noise levels above \SI{30}{dB}. Previous approaches fuse images by minimizing a convex objective function \cite{simoes2015convex} containing two quadratic data-fitting terms and an edge-preserving regularizer. The data fitting terms account for sampling rate, additive noise, and blur induced by the diffraction limit. The regularizer is a form of vector total variation (TV) which promotes a piece-wise smooth solution with discontinuities aligned across bands. TV regularization has the advantage of preserving edges, but also removes textures and fine-scale detail. TV is a strong prediction/fusion method when the spatial information has limited high-frequency spatial content (sharp edges) \cite{li2015prediction}. Since high spatial resolution is a priority in many imaging applications, maintaining high-frequency is often critical.

\begin{figure}[t]
   \centering
   \begin{tabular}{@{}c@{\hspace{.5cm}}c@{}}
        \includegraphics[scale = 0.45]{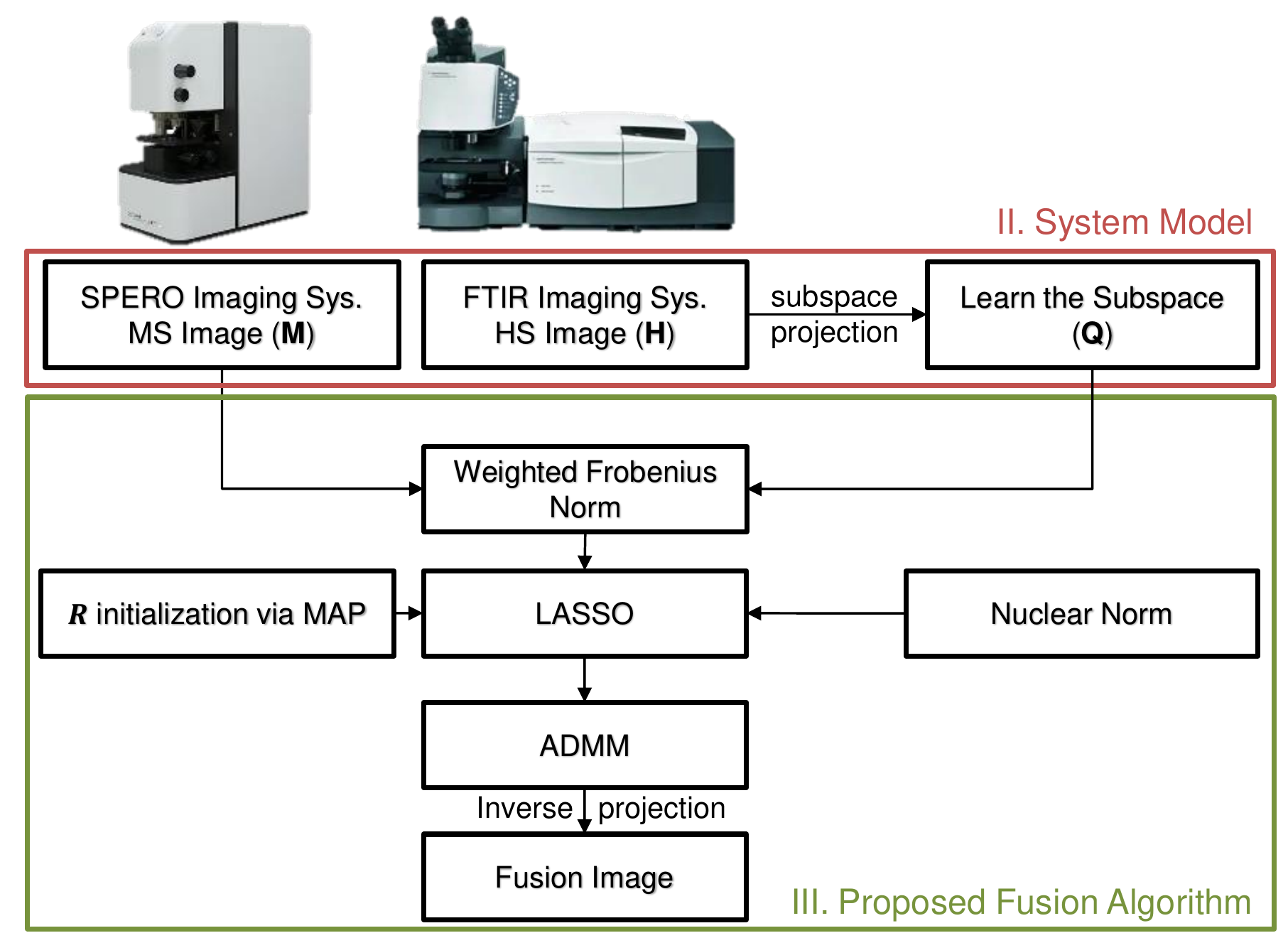}
   \end{tabular}
 \caption{Fusion Model Based on LASSO. Images from DFIR and FTIR system are used in LASSO's Frobenius norm in order to regularize and optimize the fused image.}
 \label{fig:SysModel}
\end{figure}

We propose a fusion-based approach that leverages the advantages of HSI and MSI to obtain benefits from both modalities. Our acquisition model integrates modality-specific features of the data, including noise and spectral/spatial sampling. Our proposed fusion model relies on \textit{least absolute shrinkage and selection operator} (LASSO) to regulate and optimize the fusion image. Since LASSO is costly to compute with large hyperspectral images, we employ the \textit{alternating direction method of multipliers} (ADMM) method for efficiency. Our fusion model is illustrated in Figure \ref{fig:SysModel}.

Our contributions are summarized as follows:
\begin{itemize}
    \item \textbf{Image fusion using proposed LASSO:} We use LASSO's Frobenius norm to minimize the differences between fusion and input images (MSI and HSI). LASSO's nuclear norm is used to further reduce noise and enforce sparsity in the result. 
    \item \textbf{Fused image initialization via Maximum a Posteriori (MAP)}: We initialize our fused image based on previous work by Hardie et al. in \cite{hardie2004map}.
    \item \textbf{Implementation of ADMM:} We apply ADMM decomposition to increase performance.
    \item \textbf{Performance Evaluation:} We evaluate the proposed method using a variety of fusion metrics and compare results to the state-of-the-art in noise reduction, image fusion, and pansharpening.
\end{itemize}
 
\section{System Model} \label{II}
We describe each image as a third-order tensor representing two spatial and one spectral dimension, with spectra sampled in units of wavelength $\lambda$ or wavenumber $\wn$ (\si{cm^{-1}}). The inputs have spatial pixel counts $N_H = X_H\times Y_H$ and $N_M = X_M \times Y_M$ for hyperspectral and multispectral images, respectively. We expect that $N_H\leq N_M$ since multispectral data is most useful with additional spatial detail. The spectral bands are given by $Z_H$ and $Z_M$ such that $Z_M < Z_H$. The following notation is used to express the model:
\begin{itemize}
    \item $\Hmat\in \mathbb{R}^{N_H\times Z_H}$ is the low-resolution HS input image
    \item $\Mmat\in \mathbb{R}^{N_M\times Z_M}$ is the high-resolution MS input image
    \item $\Fmat\in \mathbb{R}^{N_M\times Z_H}$ is the fused output image
    \item $s$ is the imaged object represented by a continuous function $s(\mathbf{x}, \lambda)$, where $\mathbf{x} = [x, y]^T$ is a position in space and $\lambda$ is a wavelength
    \item $\Smat \in \mathbb{R}^{N_M\times Z_H}$ is the optimal object image at the highest possible spatial and spectral resolution given the point-spread-function $p$ of the imaging system: $\Smat = p\circledast s$
\end{itemize}
The proposed algorithm uses two degraded images $\Hmat$ and $\Mmat$ of the object $s$ to reconstruct $\Fmat$ such that $\Fmat - \Smat\rightarrow \mathbf{0}$.

\subsection{Image Generation} \label{ImageGeneration}
The ideal image $\mathbf{S}$ is degraded by the imaging process. A hyperspectral image with low spatial resolution is generated by:
\begin{equation}
    \mathbf{H} = \mathbf{L}\mathbf{S} + \boldsymbol{\eta}_1,
    \label{eqn:hyperspectral}
\end{equation}
where the operator $\Lmat \in \mathbb{R}^{N_H\times N_M}$ performs blurring and resampling of $\mathbf{S}$. A multispectral image collected at high resolution is generated by:
\begin{equation}
    \Mmat = \Smat\Bmat + \boldsymbol{\eta}_2,
    \label{eqn:multispectral}
\end{equation}
where $\Bmat$ is a sparse matrix that extracts individual bands from $\Smat$. Both $\boldsymbol{\eta}_1$ and $\boldsymbol{\eta}_2$ are independent sources of additive noise.

\subsection{Principal Component Analysis} {\label{II.B}}
Subspace projection methods like principal component analysis (PCA) convert data with possibly correlated variables into a set of linearly uncorrelated values (i.e. principal components). Only the most energized (i.e. highest variance) components are kept. In HSI, noise is typically distributed isotropically throughout the spectrum, while actual signals lie within a smaller subspace. Furthermore, HS images are often large, with a \SI{1}{\milli\meter^{2}} biopsy image ($2000\times 2000 \times 371$ pixels) requiring \SI{\approx 5.8}{\giga B} at 32-bit precision. Dimension reduction reduces computational costs and memory requirements while increasing SNR by eliminating low-variance coefficients associated with noise. 

Our goal is to obtain a fusion image $\mathbf{F} \approx \mathbf{S}$, where $\mathbf{F} = [\mathbf{f}_1, ..., \mathbf{f}_{N_M}]^T$. Each row $\mathbf{f}_i = [f_{i,1}, f_{i,2}, ..., f_{i,Z_H}]$ represents the spectrum at each spatial pixel location. We represent $\mathbf{f}_i$ in the subspace $\mathbf{Q}$ as:
\begin{equation}
    \mathbf{f}_i = \mathbf{r}_i\mathbf{Q},
\end{equation}
where $\mathbf{r}_i \in \mathbb{R}^{1 \times \tilde{Z}}$ is the projection of $\mathbf{f}_i$ onto the orthonormal subspace spanned by the columns of $\mathbf{Q} \in \mathbb{R}^{\tilde{Z} \times Z_H}$ such that $\tilde{Z} \ll Z_H$. The image $\mathbf{R} \in \mathbb{R}^{N_M \times \tilde{Z}}$ projected onto this subspace is composed of the reduced spectra: $\mathbf{R} = [\mathbf{r}_1, ..., \mathbf{r}_{N_M}]^T$. Integrating the dimension reduction transformation $\Qmat$ into (\ref{eqn:hyperspectral}) and (\ref{eqn:multispectral}) yields:
\begin{equation}
    \mathbf{H} = \mathbf{L}\mathbf{R}\mathbf{Q} + \mathbf{\eta}_1,
    \label{eqn:subspaceHS}
\end{equation}
\begin{equation}
    \mathbf{M} = \mathbf{R}\mathbf{Q}\mathbf{B} + \mathbf{\eta}_2.
    \label{eqn:subspaceMS}
\end{equation}

\section{Proposed Fusion Algorithm} \label{III}
From equation (\ref{eqn:subspaceHS}) and (\ref{eqn:subspaceMS}), we propose using the Least Absolute Shrinkage and Selection Operator (LASSO) to optimize $\Rmat$. The fusion image is initialized with Maximum a Posteriori (MAP) \cite{hardie2004map} and the optimization is broken down into subproblems using the Alternating Direction Method of Multipliers (ADMM). The LASSO Frobenius norm minimizes the differences between the fusion image and accessible components of the observed images, while the nuclear norm further reduces noise and enforces sparsity as needed.

\subsection{Regularization}
A general LASSO optimization proposed by Tibshirani et al. \cite{tibshirani1996regression} is given by:
\begin{equation}
    \argmin_{\mathbf{x}} \frac{1}{2}\lVert(\mathbf{Ax - b})\rvert_F^2 + \eta \lVert\mathbf{x}\rVert_n,
\end{equation}
where
\begin{itemize}
    \item $\mathbf{A}\in \mathbb{R}^{n\times n}$
    \item $\mathbf{x}\in \mathbb{R}^{n\times 1}$ is the optimized term
    \item $\mathbf{b}\in \mathbb{R}^{n\times 1}$ is the observed data
    \item $\eta$ is a parameter that controls the degree of shrinkage
\end{itemize}

We optimize $\Rmat$ based on its relationships with the available HSI and MSI in equations (\ref{eqn:subspaceHS}) and (\ref{eqn:subspaceMS}) using the Frobenius norm:
\begin{equation}
    \argmin_{\mathbf{R}} \frac{1}{2} \left\lVert(\mathbf{H} - \mathbf{LRQ})\right\rVert_F^2+
    \frac{1}{2}\left\lVert(\mathbf{M} - \mathbf{RQB})\right\rVert_F^2.
    \label{eqn:L2}
\end{equation}

The HSI has lower resolution and SNR, therefore we implement weights for each term in Equation \ref{eqn:L2} to prioritize appropriate information from $\Mmat$. These weights are based on SNR, resolution, and blurriness. The MSI and HSI weights are on a logarithmic scale ranging from 30 to 50 and 5 to 30, respectively. After applying the weights, the equation (\ref{eqn:L2}) becomes:
\begin{multline}
    \argmin_{\mathbf{R}} \frac{1}{2} \left\lVert\mathbf{\Lambda_H}^{-\frac{1}{2}}(\mathbf{H} - \mathbf{LRQ})\right\rVert_F^2\\
    + \frac{1}{2}\left\lVert\mathbf{\Lambda_M}^{-\frac{1}{2}}(\mathbf{M} - \mathbf{RQB})\right\rVert_F^2.
    \label{eqn:LASSOL2}
\end{multline}
Equation (\ref{eqn:LASSOL2}) generates a fusion image that maximizes the integration of information from both $\mathbf{H}$ and $\mathbf{M}$. The resulting fusion image will also contain a portion of the noise from the two source images. The variable selection component of LASSO, represented by the nuclear norm, is employed to remove these noise terms:
\begin{multline}
    \argmin_{\mathbf{R}} \frac{1}{2} \left\lVert\mathbf{\Lambda_H}^{-\frac{1}{2}}(\mathbf{H} - \mathbf{LRQ})\right\rVert_F^2\\
    + \frac{1}{2}\left\lVert\mathbf{\Lambda_M}^{-\frac{1}{2}}(\mathbf{M} - \mathbf{RQB})\right\rVert_F^2
    + \eta\left\lVert\mathbf{R}\right\rVert_n.
    \label{eqn:MaxaP}
\end{multline}
The nuclear norm term minimizes the sum of magnitudes of $\Rmat$ (Section \ref{III.C}). Pixels with absolute magnitudes smaller than a threshold $T$, defined in Section \ref{III.C} as $\frac{\eta}{\mu}$, are set to zero.

\subsection{Initialization of $\mathbf{R}$} \label{III.B}
Based on work by Hardie et al. \cite{hardie2004map}, we initialize $\mathbf{\bar{R}}$ as the MAP of $\mathbf{R}$ given $\mathbf{H}$ and $\mathbf{M}$:
\begin{equation}
\begin{split}
    p(\mathbf{{\bar{R}|H,M}}) 
    &=\frac{p(\mathbf{H}|\mathbf{{\bar{R}}})p(\mathbf{\bar{R}}|\mathbf{{M}})p(\mathbf{{M}})}{p(\mathbf{H, M})}.\\
\end{split}
\label{eqn:probbarR}
\end{equation}
Since $P(\mathbf{H, M})$ is not a function of $\mathbf{\bar{R}}$, the optimization for $P(\mathbf{\bar{R}|H, M})$ (Equation \ref{eqn:probbarR}) reduces to:
\begin{align}
    \mathbf{\bar{R}} = \argmax_{\mathbf{\bar{R}}} \big[p(\mathbf{H|\bar{R}})p(\mathbf{\bar{R}|M})\big].
    \label{eqn:RbarMax}
\end{align}
Continuing to follow Hardie's work, we obtain the initialization of $\mathbf{\bar{R}}$ as: 
\begin{multline}
    \argmin_{\mathbf{\bar{R}}} {\frac{1}{2} \left\lVert\mathbf{\Lambda_H}^{-\frac{1}{2}}(\mathbf{H} - \mathbf{L\bar{R}Q})\right\rVert_F^2} \\
    + {\frac{1}{2}\left\lVert\mathbf{\Lambda_{\bar{R}|M}}^{-\frac{1}{2}}(\mathbf{\bar{R}} - \mathbf{\tilde{R}})\right\rVert_F^2},
    \label{eqn:MaxaPRbar}
\end{multline}
where
\begin{itemize}
    \item $\mathbf{\tilde{R}}$ $= \mathbb{E}\{\mathbf{\bar{R}}|\mathbf{M}\}$ is the expected value of $\mathbf{\bar{R}}$ given $\mathbf{M}$,
        \item Rewrite as: {$\mathbf{\tilde{R}}$} $= \mathbb{E}(\mathbf{\bar{R}}) + \frac{\mathbf{\Lambda}_{\mathbf{\bar{R}},\mathbf{M}} [\mathbf{M} - \mathbb{E}(\mathbf{M})]}{\mathbf{\Lambda}_{\mathbf{M,M}}}$,
            \item $\mathbf{\Lambda}_{\mathbf{\bar{R}},\mathbf{M}}$ is the cross-covariance matrices with form:\\ $\mathbf{\Lambda}_{\mathbf{\bar{R}},\mathbf{M}} =\mathbb{E}\big[ \big(\mathbf{\bar{R}} - \mathbb{E}(\mathbf{M})\big)\big(\mathbf{\bar{R}} - \mathbb{E}(\mathbf{M})\big)^T\big],$
    \item $\mathbf{\Lambda}_{\mathbf{\bar{R}}|\mathbf{M}}$ is the row covariance matrix of $\mathbf{R}$ given $\mathbf{M}$ as:
    $\mathbf{\Lambda}_{\mathbf{\bar{R}}|\mathbf{M}} = \mathbf{\Lambda}_{\mathbf{\bar{R}},\mathbf{\bar{R}}} - \frac{\mathbf{\Lambda}_{\mathbf{\bar{R}},\mathbf{M}}\mathbf{\Lambda}^T_{\mathbf{\bar{R}},\mathbf{M}}}{\mathbf{\Lambda}_{\mathbf{M},\mathbf{M}}}$.
\end{itemize}
Solving (\ref{eqn:MaxaPRbar}) directly gives us $\mathbf{\bar{R}}$:
\begin{multline}
    \mathbf{\bar{R}} =
    \Big(\mathbf{Q}^T\mathbf{L}^T\mathbf{\Lambda_H}\mathbf{L}\mathbf{Q}+ \mathbf{\Lambda_{\bar{R}|M}}\Big)^{-1}\\
    \Big(\mathbf{L}^T\mathbf{\Lambda_H}\mathbf{Q}^T\mathbf{H} + \mathbf{\Lambda_{\bar{R}|M}}^{-1}\mathbf{\tilde{R}}\Big).
    \label{eqn:RbarMAP}
\end{multline}

\subsection{Alternating Direction Method of Multipliers (ADMM)} \label{III.C}
Since hyperspectral images tend to be large, with our application data in the range of ($1128\times 1152 \times 371$), ADMM \cite{boyd2011alternating} is used to partition the optimization. ADMM is widely popular due to its ability to handle large data effectively \cite{7808985, parikh2014block, han2017signal} by breaking an expensive optimizations into less costly sub-problems. To solve the LASSO equation (\ref{eqn:MaxaP}), we introduce the splitting variables $\mathbf{W}_1 = \mathbf{LR}$ and scaled Lagrange multipliers $\mathbf{J}_1$ for the first Frobenius norm term, $\mathbf{W}_2 = \mathbf{R}$ and $\mathbf{J}_2$ for the second Frobenius norm term, $\mathbf{W}_3 = \mathbf{R}$ and $\mathbf{J}_3$ for the Nuclear norm term. The augmented Lagrangian associated with the optimization of $\mathbf{R}$ is:
\begin{multline}
   \mathcal{L}(\mathbf{R, W_1, W_2, W_3, J_1, J_2, J_3}) \\
   = \frac{1}{2}\lVert\mathbf{\Lambda}_H^{-\frac{1}{2}}(\mathbf{H} - \mathbf{W_1Q})\rVert^2_F + \frac{\mu}{2}\lVert\mathbf{LR} - \mathbf{W_1} - \mathbf{J_1}\rVert^2_F\\
   + \frac{1}{2}\lVert\mathbf{\Lambda}_M^{-\frac{1}{2}}(\mathbf{M} - \mathbf{W_2QB})\rVert_F^2 + \frac{\mu}{2}\lVert\mathbf{R} - \mathbf{W_2} - \mathbf{J_2}\rVert_F^2\\
   +\eta\lVert\mathbf{W}_3 \rVert_n + \frac{\mu}{2}\lVert\mathbf{W}_3 - \mathbf{R} -\mathbf{J}_3\rVert^2_F,
   \label{eqn:LagEq}
\end{multline}
where $\mu$ and $\eta$ are the scalar regularization parameters, with $\eta \approx 1.25\cdot 10^{-3} \cdot \lVert\mathbf{H}\rVert_{\infty}$ and $\mu \approx 5\cdot 10^{-2} \cdot \frac{\mathbf{\Lambda_H}}{Z_H}$. Minimizing (\ref{eqn:LagEq}) with respect to one term while holding others fixed allows the following update functions:
\begin{multline}
    \mathbf{R}^{t+1} = (\mathbf{LL}^T + 2\mathbf{I})^{-1} \Big[(\mathbf{W}_1^{t} + \mathbf{J}_1^{t})\mathbf{L}^T \\
    + (\mathbf{W}_2^{t} +\mathbf{J}_2^{t} + \mathbf{W}_3^{t} - \mathbf{J}_3^{t})\Big],
    \label{eqn:UpdateR}
\end{multline}
\begin{multline}
    \mathbf{W}_1^{t+1} =
    \big(\mathbf{Q}^T\mathbf{\Lambda}_H^{-1}\mathbf{Q} + \mu \mathbf{I}\big)^{-1}\\\big(\mathbf{Q}^T\mathbf{\Lambda}_H^{-1}\mathbf{H} +\mu(\mathbf{L}\mathbf{R}^{t+1}-\mathbf{J}_1^{t})\big),
    \label{eqn:UpdateW1}
\end{multline}
\begin{multline}
    \mathbf{W}_2^{t+1} =
    \big(\mathbf{B}^T\mathbf{Q}^T\mathbf{\Lambda}_M^{-1}\mathbf{Q}\mathbf{B} + \mu \mathbf{I}\big)^{-1}\\ \big(\mathbf{B}^T\mathbf{Q}^T\mathbf{\Lambda}_M^{-1}\mathbf{M} +\mu(\mathbf{R}^{t+1}-\mathbf{J}_2^{t})\big).
    \label{eqn:UpdateW2}
\end{multline}
For $\mathbf{W}_3$, we implemented soft-thresholding to solve it fast and efficiently, thus we first define the following term for the Frobenius term:
\begin{equation}
     (\mathbf{W}^{u+1}_3)^{LS} = \mathbf{R}^{u+1}+\mathbf{J}_3^{u}.
     \label{eqn:UpdateW3LS}
\end{equation}
Then, the optimization for $\mathbf{W}_3$ becomes:
\begin{equation}
    \eta\lVert\mathbf{W}^{u+1}_3 \rVert_n + \frac{\mu}{2}\lVert\mathbf{W}^{u+1}_3 - (\mathbf{W}^{u+1}_3)^{LS}\rVert^2_F,
\end{equation}
which has an optimality condition at:
\begin{multline}
    0\in \nabla(\mu\lVert\mathbf{W}^{u+1}_3 - (\mathbf{W}^{u+1}_3)^{LS}\rVert^2_F) + \partial (\eta\lVert\mathbf{W}^{u+1}_3 \rVert_n)\\
    \Leftrightarrow 0\in \mu\Big(\mathbf{W}^{u+1}_3 - (\mathbf{W}^{u+1}_3)^{LS}\Big) + \eta\partial (\lVert\mathbf{W}^{u+1}_3 \rVert_n).
    \label{eqn:W3L2L1}
\end{multline}
At pixel location $[i,j]$, where ${w}^{u+1}_{3_{i,j}} \neq 0$, we have:
\begin{equation}
    \partial (\lVert{w}^{u+1}_{3_{i,j}} \rVert_n) = sign({w}^{u+1}_{3_{i,j}}).
\end{equation}
Equation (\ref{eqn:W3L2L1}) for those pixels is re-written as:
\begin{multline}
    0\in \mu\Big({w}^{u+1}_{3_{i,j}} - ({w}^{u+1}_{3_{i,j}})^{LS}\Big) + \eta sign({w}^{u+1}_{3_{i,j}})\\
    \Leftrightarrow {w}^{u+1}_{3_{i,j}} = ({w}^{u+1}_{3_{i,j}})^{LS} - \frac{\eta}{\mu} sign({w}^{u+1}_{3_{i,j}}).
    \label{eqn:W3L2sign}
\end{multline}
In equation (\ref{eqn:W3L2sign}), we see that when ${w}^{u+1}_{3_{i,j}} < 0$, then $({w}^{u+1}_{3_{i,j}})^{LS} < - \frac{\eta}{\mu} < 0$, and ${w}^{u+1}_{3_{i,j}} > 0$ leads to $({w}^{u+1}_{3_{i,j}})^{LS} > \frac{\eta}{\mu} > 0$. Thus, $\left|({w}^{u+1}_{3_{i,j}})^{LS}\right| > 0$ and $sign({w}^{u+1}_{3_{i,j}}) = sign\big({w}^{u+1}_{3_{i,j}})^{LS}\big)$. We then rewrite (\ref{eqn:W3L2sign}) as:
\begin{multline}
    {w}^{u+1}_{3_{i,j}} = ({w}^{u+1}_{3_{i,j}})^{LS} - \frac{\eta}{\mu} sign(({w}^{u+1}_{3_{i,j}})^{LS})\\
    \Leftrightarrow w_{3_{i,j}}^{u+1} = \text{sign}\left[({w}^{u+1}_{3_{i,j}})^{LS}\right]\left(\left|({w}^{u+1}_{3_{i,j}})^{LS}\right| - \frac{\eta}{\mu}\right).
\end{multline}
If the pixel at location $[i,j]$ has ${w}^{u+1}_{3_{i,j}} = 0$, the sub-differential of the nuclear norm is the interval $[-1,1]$. We rewrite equation (\ref{eqn:W3L2L1}) for those pixels as:
\begin{multline}
    0\in -({w}^{u+1}_{3_{i,j}})^{LS} + \frac{\eta}{\mu}[-1,1]
    \Leftrightarrow \left|({w}^{u+1}_{3_{i,j}})^{LS}\right|\leq \frac{\eta}{\mu}.
\end{multline}
So we update $\mathbf{W}^{u+1}_3$ element-wise as:
\begin{multline}
    {w}^{u+1}_{3_{i,j}} =\\
    \begin{cases}
      0&\text{,$\left|({w}^{u+1}_{3_{i,j}})^{LS}\right|\leq \frac{\eta}{\mu}$}\\
      \text{$sign$}\left[({w}^{u+1}_{3_{i,j}})^{LS}\right]\left(\left|({w}^{u+1}_{3_{i,j}})^{LS}\right| - \frac{\eta}{\mu}\right)&\text{,$\left|({w}^{u+1}_{3_{i,j}})^{LS}\right|> \frac{\eta}{\mu}$}\\
    \end{cases} 
    \label{eqn:UpdateW3}
\end{multline}
$\mathbf{J}_1$, $\mathbf{J}_2$, $\mathbf{J}_3$, are calculated using:
\begin{equation}
    \mathbf{J}_1^{t+1} = \mathbf{J}_1^{t} - \mathbf{R}^{t+1}\mathbf{L} + \mathbf{W}_1^{t+1},
    \label{eqn:UpdateJ1}
\end{equation}
\begin{equation}
    \mathbf{J}_2^{t+1} = \mathbf{J}_2^{t} - \mathbf{R}^{t+1} + \mathbf{W}_2^{t+1},
    \label{eqn:UpdateJ2}
\end{equation}
\begin{equation}
    \mathbf{J}_3^{u+1} = \mathbf{J}_3^{u} - \mathbf{R}^{u+1} + \mathbf{W}_3^{u+1}.
    \label{eqn:UpdateJ3}
\end{equation}
The final algorithm is given in Algorithm \ref{alg:LASSOAlg} with ADMM variables.

\begin{algorithm}
    \SetAlgoLined
    \SetKwInOut{Input}{Input}
    \SetKwInOut{Output}{Output}
    \Input{$ \mathbf{H, M, \Lambda_H, \Lambda_M}, \tilde{Z}, \mathbf{B, L,} n_{it}, \mu, \eta, \gamma$}
    Compute $\mathbf{\bar{R}}$\\
    $\mathbf{Q} \leftarrow PCA(\mathbf{H}, \tilde{Z})$ \%Identify the HS image subspace\\
    \For{$u = 1$ to $ite_{L1}$}{
        \%---------------- L2 Regularization ----------------\%\\
        \For{$t = 1$ to $ite_{L2}$}{
            \%Optimize $\mathbf{R}$ w. ADMM\\
            $\mathbf{R}^{t+1} = (\mathbf{LL}^T + 2\mathbf{I})^{-1} \Big[(\mathbf{W}_1^{t} + \mathbf{J}_1^{t})\mathbf{L}^T + (\mathbf{W}_2^{t} +\mathbf{J}_2^{t} + \mathbf{W}_3^{t} - \mathbf{J}_3^{t})\Big]$ \Big(Eqn. (\ref{eqn:UpdateR})\Big)\\
            \%Update $\mathbf{W}_1$\\
            $\mathbf{W}_1^{t+1} =
            \big(\mathbf{Q}^T\mathbf{\Lambda^{-1}_H}\mathbf{Q} + \mu \mathbf{I}\big)^{-1}\big(\mathbf{Q}^T\mathbf{\Lambda^{-1}_H}\mathbf{H} +\mu(\mathbf{R}^{t+1}\mathbf{L}-\mathbf{J}_1^{t})\big)$ \Big(Eqn. (\ref{eqn:UpdateW1})\Big)\\
            \%Update $\mathbf{J}_1$\\
            $\mathbf{J}_1^{t+1} = \mathbf{J}^{t}_1-\mathbf{R}^{t+1}+\mathbf{W}_1^{t+1}$\\
            \%Update $\mathbf{W}_2$ \Big(Eqn. (\ref{eqn:UpdateJ1})\Big)\\
            $\mathbf{W}_2^{t+1} =
            \big(\mathbf{B}^T\mathbf{Q}^T\mathbf{\Lambda^{-1}_M}\mathbf{Q}\mathbf{B} + \mu \mathbf{I}\big)^{-1} \big(\mathbf{B}^T\mathbf{Q}^T\mathbf{\Lambda^{-1}_M}\mathbf{M} +\mu(\mathbf{R}^{t+1}-\mathbf{J}_2^{t})\big)$ \Big(Eqn. (\ref{eqn:UpdateW2})\Big)\\
            \%Update $\mathbf{J}_2$\\
            $\mathbf{J}_2^{t+1} = \mathbf{J}_2^t-\mathbf{R}^{t+1}+\mathbf{W}_2^{t+1}$ \Big(Eqn. (\ref{eqn:UpdateJ2})\Big)\\
        }
        \%---------------- L1 Regularization ----------------\%\\
        $\mathbf{R}^{u+1} = \mathbf{R}^{t+1}$\\
        \%Update each pixel in $\mathbf{W}_3$\\
        $(\mathbf{W}^{u+1}_3)^{LS} = \mathbf{R}^{u+1} - \mathbf{J}_3^{u}$ \Big(Eqn. (\ref{eqn:UpdateW3LS})\Big)\\
        $w_{3_{i,j}}^{u+1} = sgn\big(({w}^{u+1}_{3_{i,j}})^{LS}\big)(\left|({w}^{u+1}_{3_{i,j}})^{LS}\right| - \frac{\eta}{\mu})^+$ \Big(Eqn. (\ref{eqn:UpdateW3})\Big)\\
        \%Update $\mathbf{J}_3$\\
        $\mathbf{J}_3^{u+1} = \mathbf{J}_3^{u}-\mathbf{R}^{u+1}+\mathbf{W}_3^{u+1}$ \Big(Eqn. (\ref{eqn:UpdateJ3})\Big)\\
    }
    Set $\mathbf{{F}} = \mathbf{{R}}^{(u+1)}\mathbf{{Q}}$\\
    \textbf{Output}:$\mathbf{{F}}$ \text{(high resolution HS image)}
    \caption{Algorithm with ADMM implemented}
    \label{alg:LASSOAlg}
\end{algorithm}

\subsection{Complexity Analysis}

Each iteration of the first $\mathbf{for}$ loop has a computational complexity of $\mathcal{O}(ite_{L2}\tilde{Z} N_{M} \log(\tilde{Z} N_{M}))$ when $\tilde{Z} \leq \log (\tilde{Z} N_M)$, and $\mathcal{O}(ite_{L2}\tilde{Z}^2 N_{M})$ otherwise.

The following terms are independent of iteration and pre-computed:
\begin{itemize}
    \item $(\mathbf{LL}^T + 2I)^{-1}$
    \item $(\mathbf{Q}^T\mathbf{\Lambda^{-1}_H}\mathbf{Q} + \mu \mathbf{I})^{-1}$
    \item $\mathbf{Q}^T\mathbf{\Lambda^{-1}_H}\mathbf{H}$
    \item $(\mathbf{B}^T\mathbf{Q}^T\mathbf{\Lambda^{-1}_M}\mathbf{Q}\mathbf{B} + \mu \mathbf{I})^{-1}$
    \item $\mathbf{B}^T\mathbf{Q}^T\mathbf{\Lambda^{-1}_M}\mathbf{M}$
\end{itemize}
In Algorithm \ref{alg:LASSOAlg}, line 7 calculates the update for $\mathbf{R}$, which is performed through Fast Fourier Transform (FFT) with a complexity $\mathcal{O}(\tilde{Z} N_{M} \log(\tilde{Z} N_{M}))$ at each iteration inside the second loop. Line 9 and 13 have a complexity of $\mathcal{O}(\tilde{Z}^2 N_{M})$. In line 11 and line 15, calculations involve only matrix additions which has the complexity of $\mathcal{O}(\tilde{Z} N_{M})$. The $L2$ component of the proposed regularization therefore exhibits complexity similar to HySure \cite{simoes2015convex} and Bayes Sparse \cite{wei2015hyperspectral}. The proposed sparse regularization in lines 20 and 23 have a complexity of $\mathcal{O}(\tilde{Z} N_{M})$. Line 21 is a soft threshold operator which has a similar complexity of $\mathcal{O}(\tilde{Z} N_{M})$. Bayes Sparse \cite{wei2015hyperspectral} has a complexity for their patch-wise sparse coding as $\mathcal{O}(K n_p n_{pat}\tilde{m}_{\lambda})$, where $K$ is the maximum number of atoms in a patch, $n_p$ is the number of patch pixels, $n_{pat}$ is the number of patches per band in the image cube (e.g. HSI, MSI), and $\tilde{m}_{\lambda}$ is the number of subspace bands. We have $\tilde{Z} = \tilde{m}_{\lambda}$, $n_{pat} \approx N_M$ (compared to Bayes Sparse's code implementation), so the speed up in our code compared to Bayes Sparse is $Kn_p$, with $K = 4$, and $n_p = 36$ based on the default values in Bayes Sparse's code implementation.

\begin{figure}
  \includegraphics[scale = 0.47]{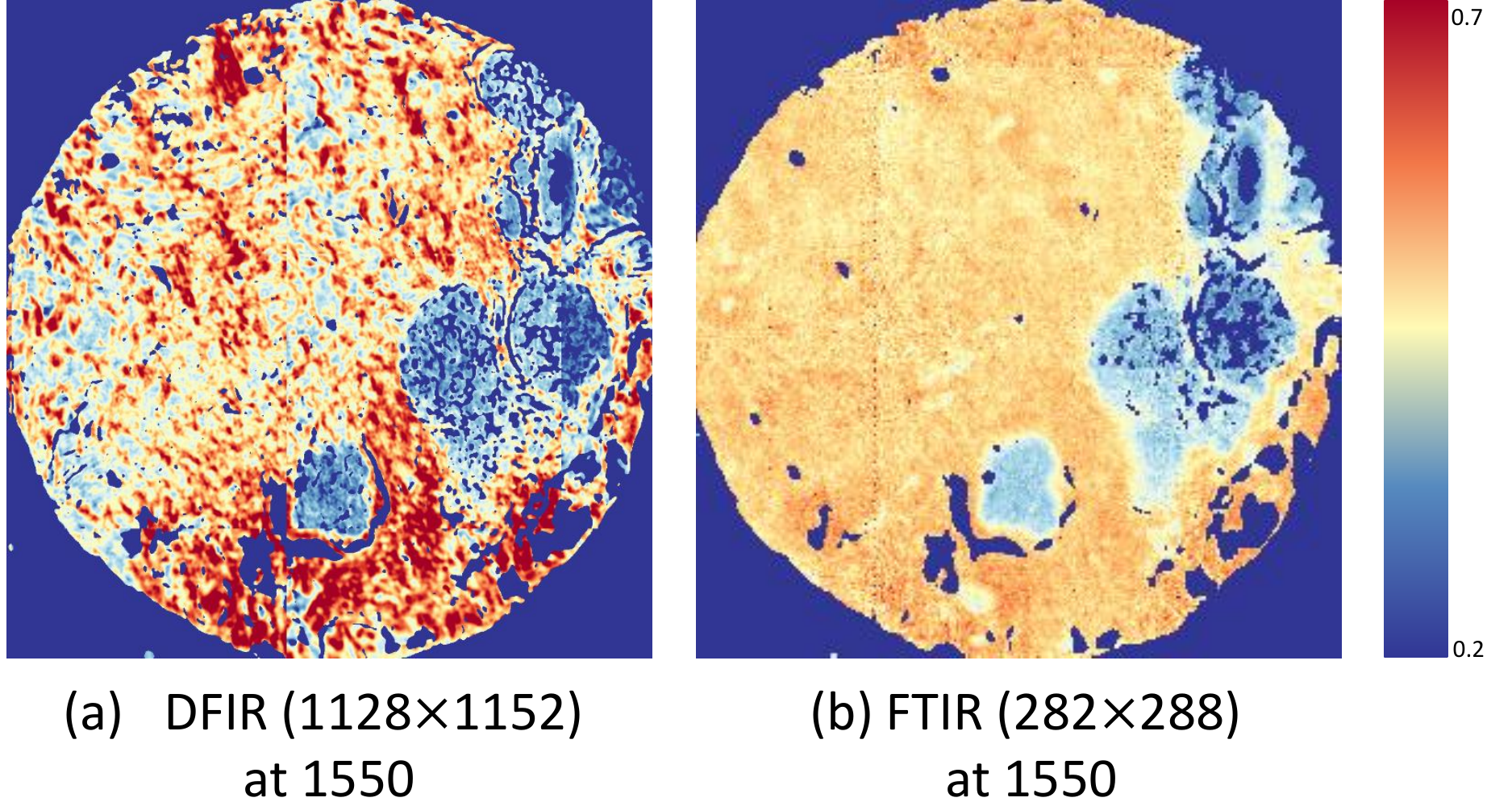}
      \caption{MSI and HSI from breast tissue taken by DFIR and FTIR imaging systems.}
  \label{fig:TissueDFIRFTIR}
\end{figure}
\begin{figure*}
  \includegraphics[width=\textwidth]{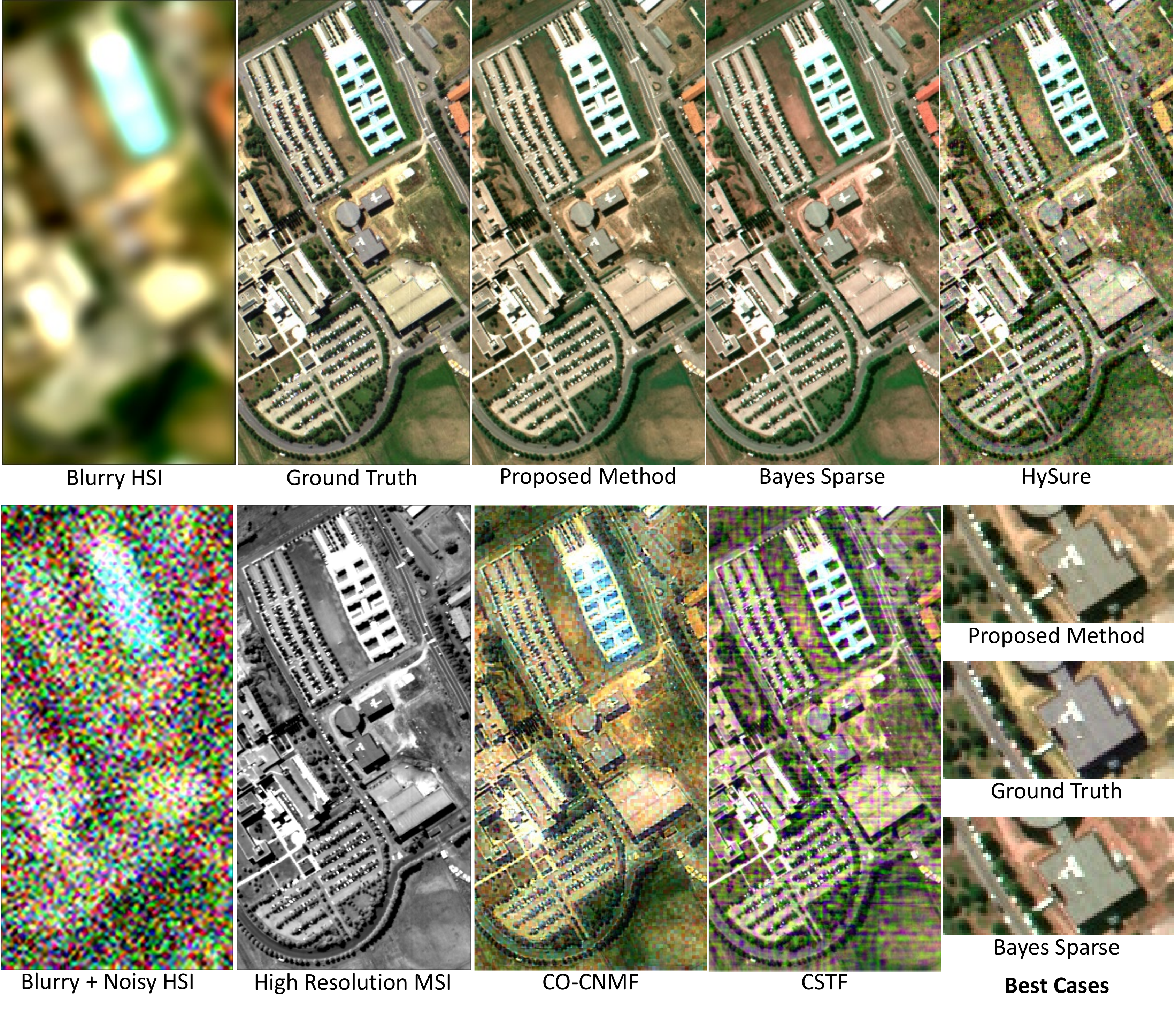}
      \caption{University of Pavia dataset. Blurry HSI was blurred with a 39x39 Gaussian kernel and then downsampled by every 4 pixels, Gaussian noise is added at 10dB SNR (HSI) and 50 dB SNR (MSI).Best Cases shows the close-up comparison between our method and Bayes Sparse.}
  \label{fig:Pavia}
\end{figure*}
\begin{figure*}
  \includegraphics[scale = 0.34]{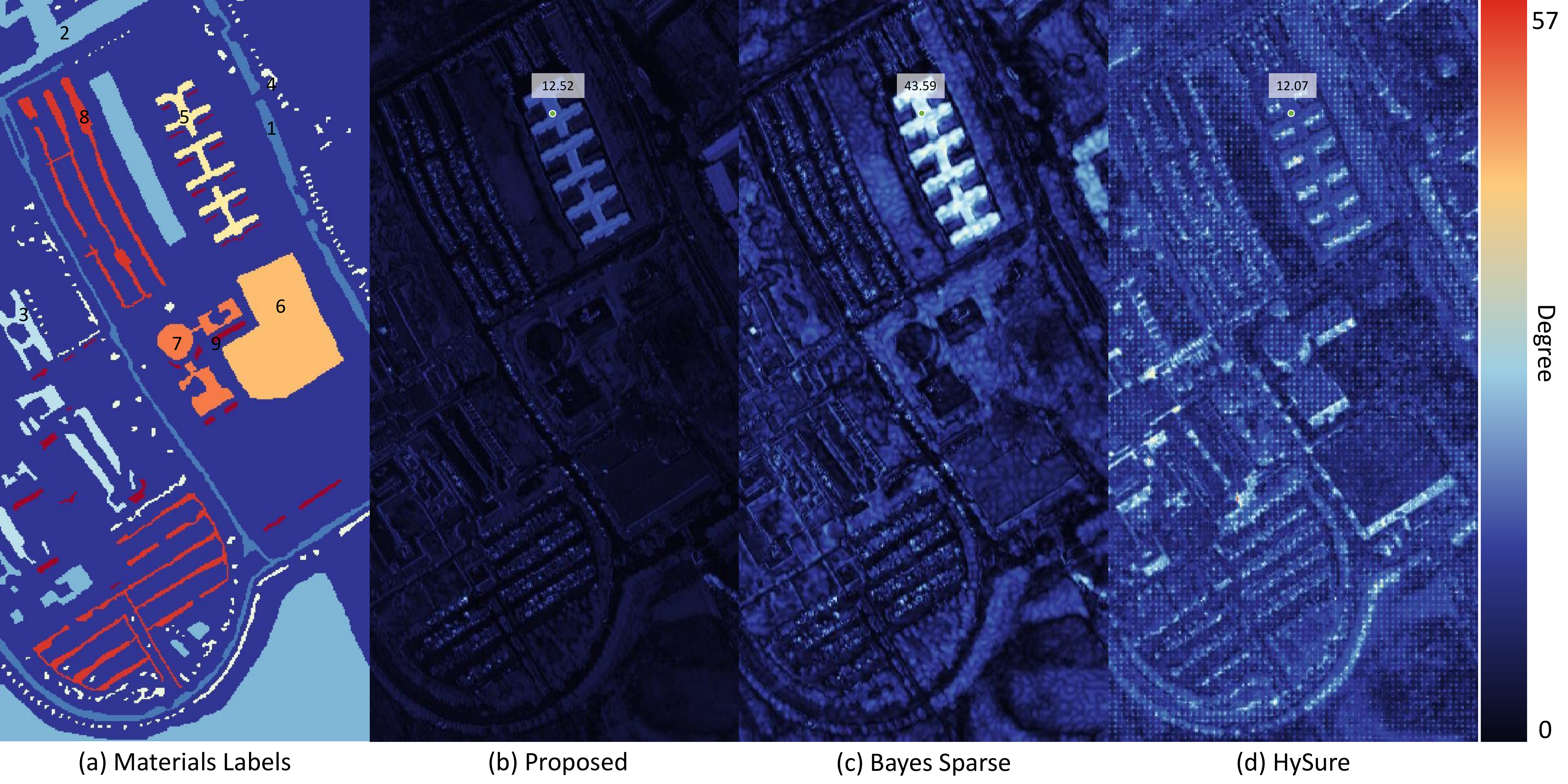}
      \caption{Pavia University's ground truth labels and spectral angle mapping (SAM) results per pixel. (a) The ground truth labels provided by expert annotation are shown for context. (b-d) SAM angle at each pixel as a result of image fusion using (b) Proposed, (c) Bayes Sparse, and (d) HySyre. Ground truth materials are annotated as: (1) asphalt, (2) meadows, (3) gravel, (4) trees, (5) painted metal sheets, (6) bare soil, (7) bitumen, (8) self-locking bricks, and (9) shadows.}
  \label{fig:SAM2}
\end{figure*}
\section{Results} \label{IV}
\subsection{Data Collection and Simulation} \label{IV.A}

In this section, we compare our algorithm to Bayes Sparse \cite{wei2015hyperspectral}, convex optimization (HySure) \cite{simoes2015convex}, coupled sparse tensor factorization (CSTF) \cite{8359412}, and convex optimization-based coupled nonnegative matrix factorization (CO-CNMF) \cite{8107710}. We first describe our image acquisition and simulation methods, including the infrared microscopy approach that motivates this work (Section \ref{IV.A}). We then describe the fusion metrics used to evaluate the resulting images (Section \ref{IV.B}). Finally, we compare our proposed fusion framework on a variety of publicly available data sets and summarize the results (Section \ref{IV.C}).
\begin{figure}
\centering
  \includegraphics[scale = 0.23]{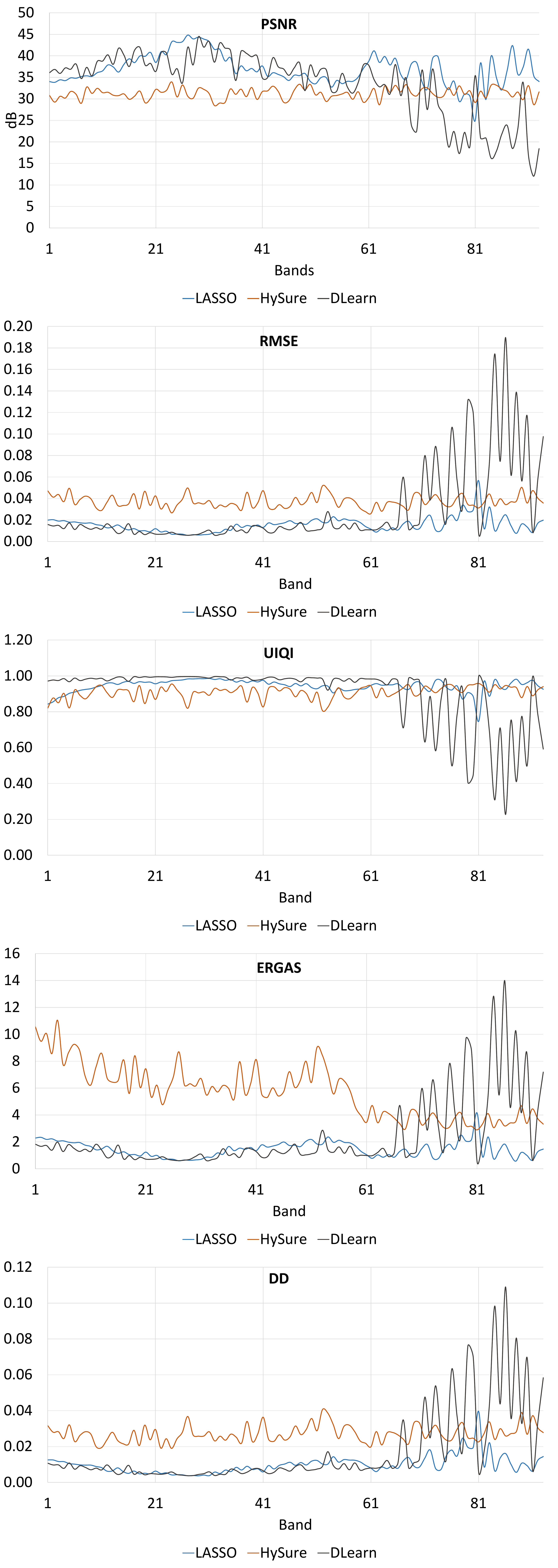}
      \caption{Metrics performance of fusion methods as a function of spectral band (Pavia University data set with Gaussian noise). Top to bottom: PSNR, RMSE, UIQI, ERGAS, DD.}
  \label{fig:PaviaUniMetrics}
\end{figure}

Arial images of the University of Pavia and Pavia Center were acquired from the Deutsches Zentrum fur Luft-und Raumfahrt under sponsorship from European Union's Hysens project. The raw images have a spatial resolution of \SI{1.3}{\meter} with 103 spectral bands. HSI data was simulated by applying a $39\times 39$ pixel Gaussian filter and downsampling every 4 pixels to reduce the spatial resolution to $\approx$\SI{5.2}{\meter}. We determine the Gaussian filter size by applying it on MSI from DFIR imaging system demonstrated in Figure \ref{fig:TissueDFIRFTIR}.a. We increase the filter size until MSI's spatial information in Collagen and Epithelium areas cannot be distinguished similar to HSI taken by FTIR imaging system in Figure  \ref{fig:TissueDFIRFTIR}.b. MSI data is simulated by randomly selecting from the first 70 bands. Gaussian noise is added at 5dB to 10dB SNR (HSI) and at 50dB SNR (MSI). In University of Pavia and Pavia Center datasets, the MSI is $512\times 256\times 4$ pixels and the HSI is $128\times 64\times 93$ pixels. The University of Pavia dataset was tested with Gaussian noise, then Pavia Center dataset with Gaussian (Gau.) noise and Poisson (Poi.) noise. 

Infrared spectroscopic images were obtained using two commercially available instruments: (1) HSI images were collected using an Agilent Cary 620 infrared spectroscopic microscope with \SI{5.5}{\micro\meter} spatial resolution and a wavelength range of \numrange{1000}{4000}\si{\centi\meter^{-1}}, and (2) MSI images were acquired using a Daylight SPERO microscope with a spatial resolution of \SI{1.3}{\micro\meter} and discrete sampling at wavelengths within a range of \numrange{900}{1700}\si{\centi\meter^{-1}}. Images of an identical breast biopsy core were collected using both systems. The HSI data was collected at high-speed using a single pass (without averaging). While traditionally designed for discrete-band imaging, the SPERO microscope was used to collect a complete spectrum from $\SI{1000}{\nano\meter}$ to $\SI{1700}{\nano\meter}$ to provide a ground truth for comparison. Individual bands were drawn from this full data cube to represent MSI input to our algorithm. In this data set, MSI has a resolution of $1128\times 1152\times 10$ pixels and HSI has a resolution of $282\times 288\times 371$ pixels. We did not calculate for SNR in HSI and MSI, nor do we calculate for the blur kernel in them. We gave them a guessed estimate value for HSI at 30dB SNR and MSI at 50db SNR. For blur kernel, we apply a Gaussian blur kernel on MSI image until they resemble HSI image. The size of the Gaussian blur kernel is 39x39.

The Indian Pines (IP) dataset is obtained by AVIRIS sensor over the Indian Pines test site in North-western Indiana, sponsored by NASA's Jet Propulsion Laboratory. The dataset has a spatial resolution of \SI{20}{m}, and 224 spectral bands. Salinas (Sal.) datasets is also obtained by AVIRIS sensor over Salinas Valley, California. It has a high spatial resolution at \SI{3.7}{m}, and 224 spectral bands. Last dataset is Kennedy Space Center (KSC), Florida which is also taken by AVIRIS sensor. It has a spatial resolution of \SI{18}{m} and 224 spectral bands. These three datasets are applied a similar blur filter and downsampling factor to Pavia University and Pavia Center. We then added them with Gaussian and Poisson noise in each experiment at \SI{10}{dB} SNR. In Indian Pines dataset, the MSI has a resolution of $128\times 128\times 10$, and HSI has a resolution of $32\times 32\times 200$. In Salinas dataset, the MSI has a resolution of $256\times 128\times 10$, and HSI has a resolution of $64\times 32\times 204$. In Kennedy Space Center dataset, the MSI has a resolution of $256\times 128\times 10$, and HSI has a resolution of $64\times 32\times 128$.

All algorithms are adjusted to be fed with the same blurred and noise added data set in simulation case. In breast tissue case, SNR level of HSI is divided into two regions, one that overlapped with MSI's spectra (guessed SNR set at $30$dB) and one that extends beyond MSI's spectra (guessed SNR set at $50$dB). MSI has guessed SNR at $50$dB. All algorithms are again provided with the same guessed SNR level of HSI and MSI. HS images are assumed to be blurry, so we also added an estimated blurred kernel to all algorithms. 

\subsection{Fusion Metrics} \label{IV.B}
Peak signal to noise ratio (PSNR) is the ratio of peak signal power to noise power expressed in term of logarithmic a decibel scale:
\begin{equation}
    \text{PSNR}(\mathbf{{F}}, \mathbf{S}) = 10\log_{10}\Bigg(\frac{\max(\mathbf{{S}})^2}{\text{MSE}(\mathbf{S}, \mathbf{{F}})}\Bigg).   
\end{equation}
where $\mathbf{S}$ is the reference image and $\mathbf{\hat{F}}$ is the fused image.

\begin{table*}[htbp]
  \centering
  \caption{HS + MS FUSION METHODS PERFORMANCES ON PAVIA UNIVERSITY DATA SET WITH GAUSSIAN NOISE (PSNR in decibels, SAM in degrees, Algorithm Time in seconds, other metrics are unitless)}
    \begin{tabular}{|r|c|c|c|c|c|c|c|c|}
    \toprule
    \multicolumn{1}{|c|}{HS Noise} & Methods & PSNR  & RMSE ($10^{-2}$) & SAM   & UIQI ($10^{-1}$) & ERGAS & DD ($10^{-2}$) & Alg. Time (s) \\
    \midrule & Bayes Sparse & 25.78 & 5.14  & 6.40  & 8.98  & 3.97  & 1.96  & 175.77 \\
\cmidrule{2-9}    \multicolumn{1}{|r|}{} & Proposed & \textbf{34.84} & \textbf{1.81} & \textbf{2.46} & \textbf{9.65} & \textbf{1.79} & \textbf{1.03} & \textbf{5.02} \\
\cmidrule{2-9}    \multicolumn{1}{|c|}{5 dB} & CO-CNMF & 23.51 & 6.68  & 14.09 & 7.62  & 9.62  & 4.90  & 73.25 \\
\cmidrule{2-9}          & CSTF  & 17.64 & 13.12 & 31.58 & 5.60  & 19.69 & 9.62  & 132.41 \\
\cmidrule{2-9}          & HySure & 18.93 & 11.30 & 29.64 & 5.10  & 35.16 & 8.50  & 39.18 \\
\cmidrule{1-9}          & Bayes Sparse & 31.42 & 2.69  & 3.80  & 9.40  & 2.30  & 1.41  & 147.79 \\
\cmidrule{2-9}    \multicolumn{1}{|r|}{} & Proposed & \textbf{35.36} & \textbf{1.71} & \textbf{2.36} & \textbf{9.68} & \textbf{1.72} & \textbf{0.99} & \textbf{5.28} \\
\cmidrule{2-9}    \multicolumn{1}{|c|}{8 dB} & CO\_CNMF & 24.77 & 5.78  & 11.76 & 8.13  & 8.15  & 4.23  & 63.92 \\
\cmidrule{2-9}          & CSTF  & 22.14 & 7.81  & 21.49 & 7.68  & 11.86 & 5.73  & 122.01 \\
\cmidrule{2-9}          & HySure & 27.21 & 4.36  & 13.22 & 8.17  & 13.12 & 3.35  & 37.95 \\
\cmidrule{1-9}          & Bayes Sparse & 31.77 & 2.58  & 4.09  & 9.44  & 2.19  & 1.48  & 143.82 \\
\cmidrule{2-9}    \multicolumn{1}{|r|}{} & Proposed & \textbf{35.53} & \textbf{1.67} & \textbf{2.32} & \textbf{9.69} & \textbf{1.69} & \textbf{0.98} & \textbf{5.18} \\
\cmidrule{2-9}    \multicolumn{1}{|c|}{10 dB} & CO\_CNMF & 25.23 & 5.48  & 10.83 & 8.27  & 7.74  & 4.00  & 67.98 \\
\cmidrule{2-9}          & CSTF  & 25.53 & 5.29  & 15.47 & 8.73  & 8.08  & 3.91  & 122.49 \\
\cmidrule{2-9}          & HySure & 30.42 & 3.01  & 9.23  & 8.93  & 8.89  & 2.29  & 37.82 \\
    \bottomrule
    \end{tabular}%
  \label{tab:PaviaTab}%
\end{table*}%

Root-mean-square error (RMSE) is used to measure the differences between the predicted model $\mathbf{F}$ to a ground truth $\mathbf{S}$. The smaller the value of RMSE, the better the fusion image quality:
\begin{equation}
    \text{RSME}(\mathbf{{F}},\mathbf{S}) = \frac{1}{N_M\times Z_H}\lVert\mathbf{{F}}-\mathbf{S}\rVert_F^2.
\end{equation}
where $N_M$ is the number of pixels in MS image, $Z_H$ is the number of spectral bands in HS image.

The \textit{spectral angle mapper} (SAM), proposed by Alparone \cite{4305345}, quantifies similarities by calculating the spectral difference between fused image and the ground truth at each spatial pixel and then averaging them. SAM of the fusion image and ground truth vector $\mathbf{{f}}_{i}$ ($\mathbf{{f}}_i = [{f}_{i,1}, {f}_{i,2}, ..., {f}_{i,Z_H}]$) and $\mathbf{s}_{i}$ ($\mathbf{s}_i = [s_{i,1}, s_{i,2}, ..., s_{i,Z_H}]$) with $i\in N_M$ are defined as:
\begin{equation}
    \text{SAM}(\mathbf{{f}}_{i}, \mathbf{s}_{i}) = \arccos{\Bigg(\frac{\langle \mathbf{{f}}_{i}, \mathbf{s}_{i} \rangle}{\lVert \mathbf{{f}}\rVert_2 \lVert\mathbf{s}_{i}\rVert_2}\Bigg)}.
\end{equation}
The final SAM value is the average across all pixels and ranges from (-90, 90] degrees. An optimal SAM value is at $0$ when the fusion image is exactly the same as ground truth.

The \textit{universal image quality index} (UIQI) was proposed by Wang and Bovik \cite{wang2002universal} to focus on the luminance, contrast, and struture of the fusion image. UIQI is calculated on each band of the two images. The UIQI between two single band images $\mathbf{s_i} = [s_{1,i}, s_{2,i}, ..., s_{N_M,i}]$ and $\mathbf{{f_i}} = [{{f}}_{1,i}, {{f}}_{2,i}, ..., {{f}}_{N_M,i}]$, $i\in Z_H$, is defined as:
\begin{equation}
\text{UIQI}(\mathbf{s},\mathbf{{f}}) = \frac{4\sigma^2_{\mathbf{s}\mathbf{{f}}}\mu_{\mathbf{s}}\mu_{\mathbf{{f}}}}{(\sigma^2_\mathbf{s} + \sigma^2_{\mathbf{{f}}})(\mu^2_\mathbf{s} + \mu^2_{\mathbf{{f}}})}    
\end{equation}
where $\sigma_{\mathbf{s},\mathbf{{f}}}$ is the covariance of ($\mathbf{s},\mathbf{{f}}$) while others values are means and variances. UIQI lies in the range [-1,1], where 1 implies that $\mathbf{{f}} = \mathbf{s}$. The comprehensive UIQI for an image is the average across all bands.
\begin{figure*}
  \centering
  \includegraphics[scale = 0.48]{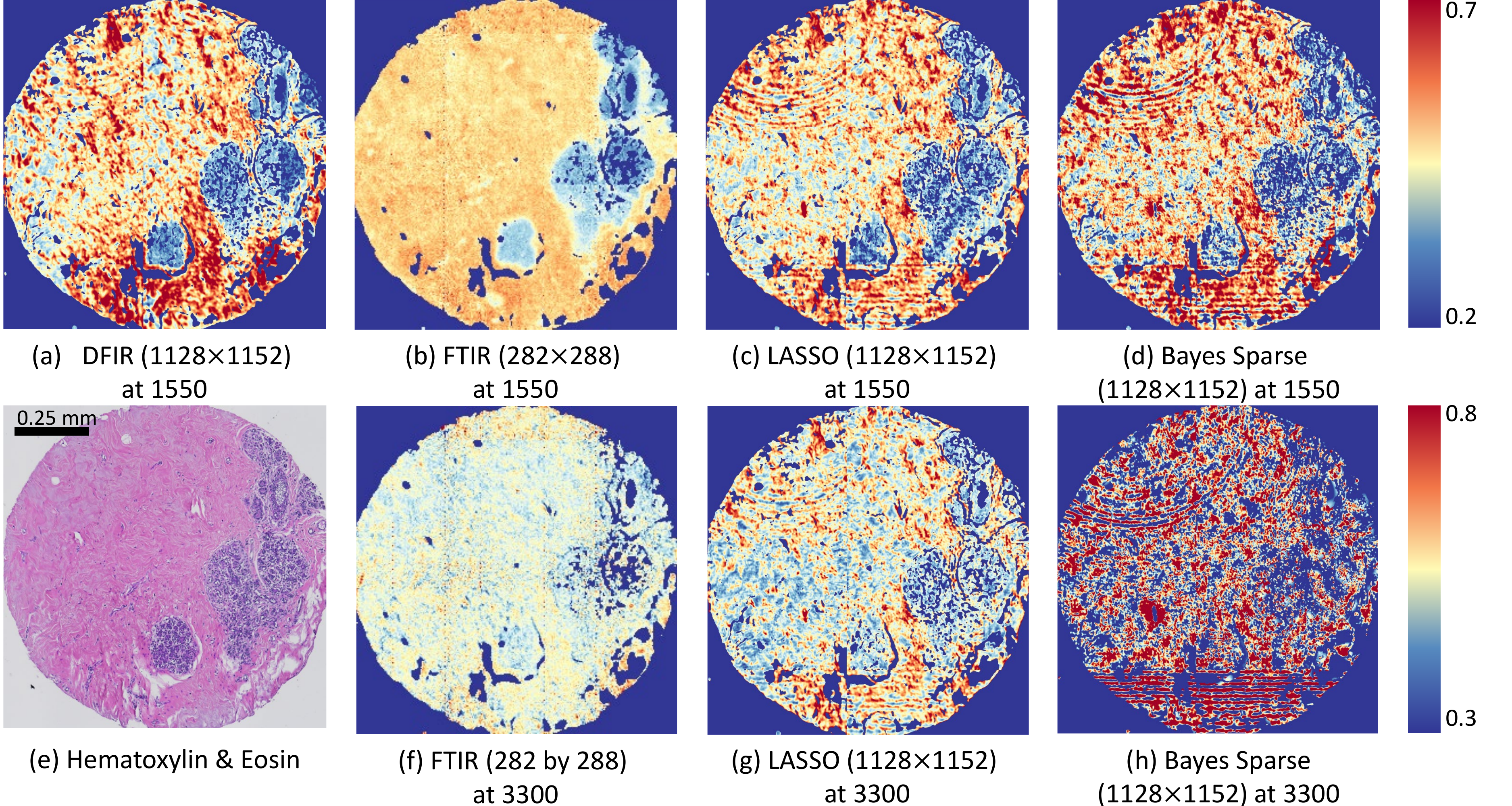}
      \caption{Infrared images of breast tissue biopsies. at \SI{1550}{\centi\meter^{-1}} (first row) and \SI{3300}{\centi\meter^{-1}} (second row) bands. Top images show the \SI{1550}{\centi\meter^{-1}} band for (a) DFIR (MS) and (b) FTIR (HS), along with fusion results using (c) the proposed LASSO method and (d) Bayes Sparse. Bottom images show (e) a traditional stained histology slide alongside the (f) FTIR image. Reconstructed results for this band are also shown using (g) Proposed and (h) Bayes Sparse. Note DFIR data is not available at \SI{3300}{\centi\meter^{-1}}.}
  \label{fig:Tissue2900}
\end{figure*}

The \textit{relative dimensionless global error in synthesis} (ERGAS) , proposed by Du \cite{4317530}, measures changes in mean and dynamic range, while also providing a comprehensive estimate of image quality. ERGAS calculates the amount of spectral distortion in the image using:
\begin{equation}
    \text{ERGAS} = 100 \times \frac{N_H}{N_M} \sqrt{\frac{1}{Z_H}\sum_{i=1}^{Z_H} \Bigg(\frac{\text{RMSE}(i)}{\mu_i}\Bigg)^2},
\end{equation}
where $\mu_i$ is the mean of the fused image at band $i$. A small ERGAS indicates less geometric or radiometric distortion, when ERGAS is $0$ it means $\mathbf{{F}} = \mathbf{S}$. 

The \textit{degree of distortion} (DD), introduced by Zhu and Bamler \cite{6319382}, which calculates the differences between two images $\mathbf{S}, \mathbf{{F}}$, is defined as: 
\begin{equation}
    DD(\mathbf{S, {F}}) = \frac{1}{N_H\times Z_H} \mid \text{vec}(\mathbf{S}) - \text{vec}(\mathbf{{F}})\mid.
\end{equation}
Note that the tensor $\mathbf{S}$ and $\mathbf{{F}}$ are vectorized in this expression. Smaller DD indicates a higher quality fusion image.
\begin{table}[htbp]
  \centering
  \caption{Tissue Data Set's Performance Result Compared to DFIR}
    \begin{tabular}{|l|r|r|}
    \toprule
    Perf  & \multicolumn{1}{l|}{Proposed} & \multicolumn{1}{l|}{Bayes Sparse} \\
    \midrule
    PSNR (dB)  & \textbf{61.41} & 57.65 \\
    \midrule
    RMSE ($10^{-2}$) & \textbf{4.58} & 5.08 \\
    \midrule
    UIQI ($10^{-1}$) & 6.70  & \textbf{6.88} \\
    \midrule
    DD ($10^{-2}$) & \textbf{2.31} & 2.56 \\
    \midrule
    Fusion Time (s) & \textbf{354.2} & 2170.9 \\
    \midrule
    Alg. Time (s) & \textbf{479.3} & 6976.9 \\
    \bottomrule
    \end{tabular}%
  \label{tab:TissuePerf2}%
\end{table}%

\begin{table}[htbp]
  \centering
  \caption{Tissue Data Set's Performance Result Compared to FTIR}
    \begin{tabular}{|l|c|c|}
    \toprule
    Perf. Metrics & \multicolumn{1}{l|}{Proposed} & \multicolumn{1}{l|}{Bayes Sparse} \\
    \midrule
    PSNR (dB) & \textbf{23.35} & 18.95 \\
    \midrule
    RMSE ($10^{-2}$) & \textbf{6.80} & 11.30 \\
    \midrule
    UIQI ($10^{-1}$) & \textbf{3.06}  & {2.23} \\
    \midrule
    DD ($10^{-2}$) & \textbf{2.71} & 5.10 \\
    \midrule
    Fusion Time (s) & \textbf{354.2} & 2170.9 \\
    \midrule
    Total Time (s) & \textbf{479.3} & 6976.9 \\
    \bottomrule
    \end{tabular}%
  \label{tab:TissuePerf1}%
\end{table}%
\begin{figure*}
    \centering
    \includegraphics[width=\textwidth]{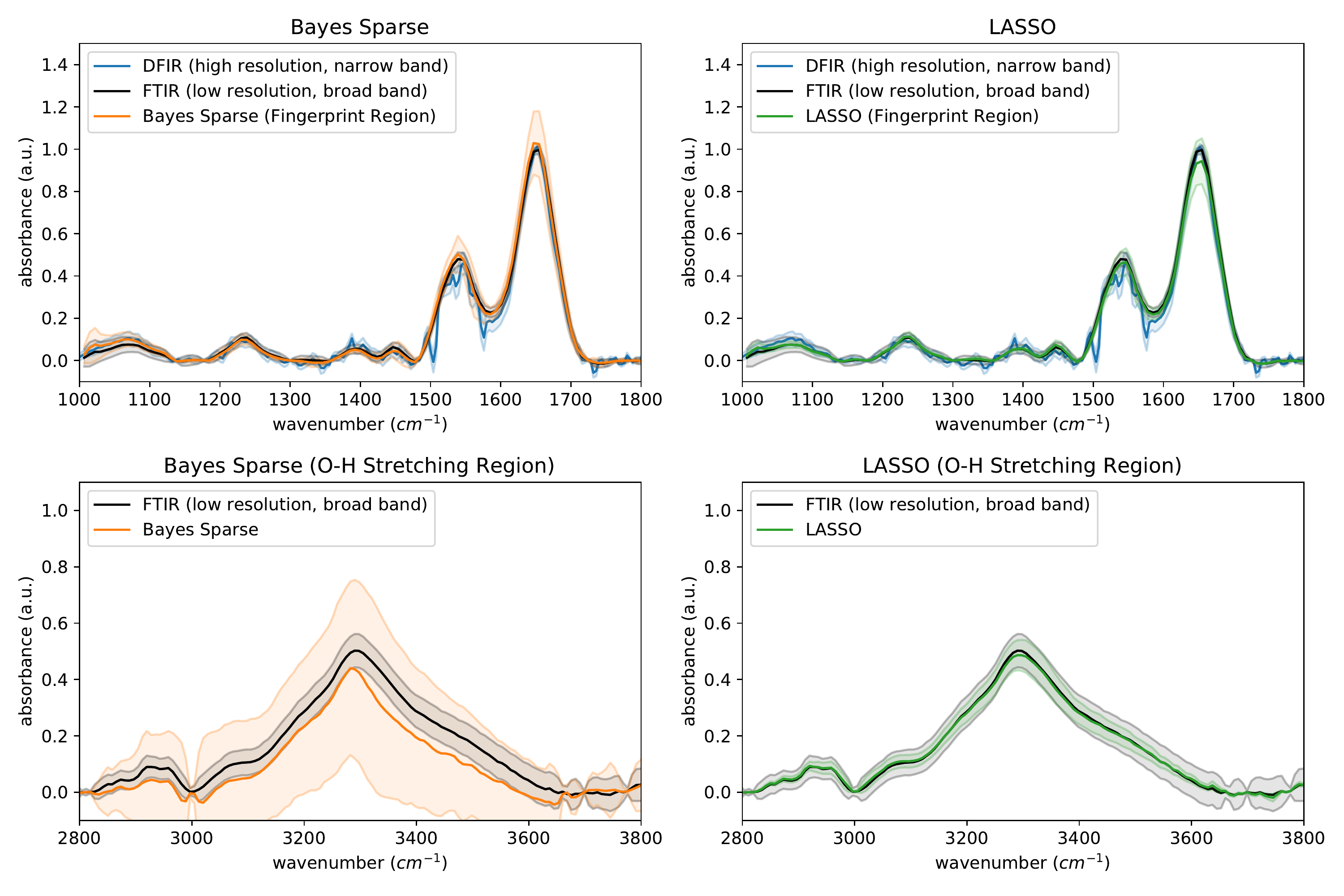}
    \caption{Spectral response for pixels in the \textit{epithelium} cell class across four images of a breast tissue: DFIR, FTIR, Dictionary Learning, LASSO images. The spectral resolution is 371 bands spanning from $1000 cm^{-1}$ to $3700 cm^{-1}$, except DFIR which has 220 bands spanning from $1000 cm^{-1}$ to $1760 cm^{-1}$. In this graph, the solid line is the mean of an area in the epithelium region of the tissue. The shaded area is the standard deviation of the pixels in the area. The blue graph depicts the DFIR image's spectral response, the grey graph depicts the FTIR image's spectral response, the orange graph depict dictionary learning image's spectral response, the green graph depict LASSO image's spectral response.}
    \label{fig:Epithelium1}
\end{figure*}
\begin{figure*}
    \centering
    \includegraphics[width=\textwidth]{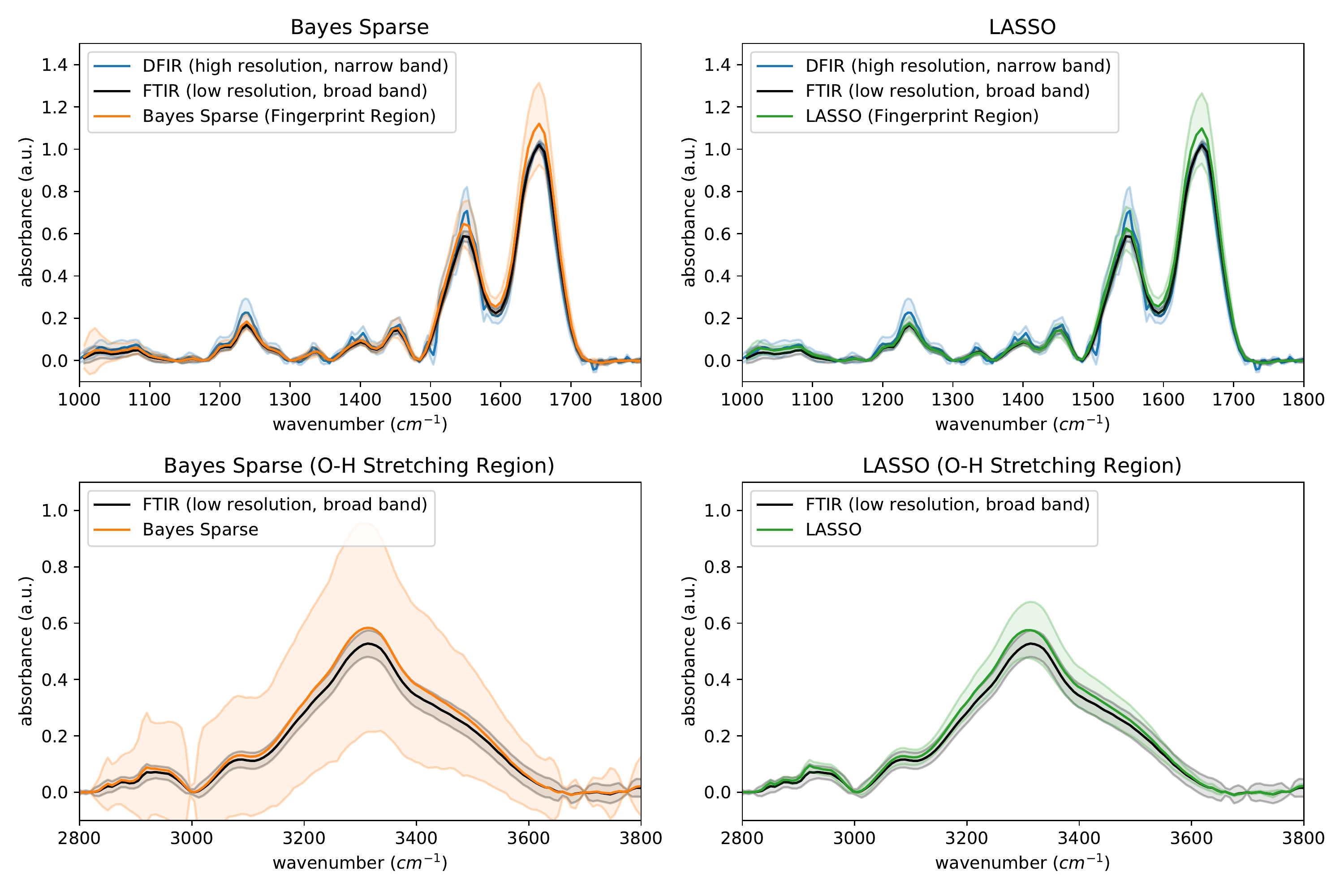}
    \caption{Spectral response for pixels in the \textit{Collagen} cell class across four images of a breast tissue: DFIR, FTIR, Dictionary Learning, LASSO images. The spectral resolution is 371 bands spanning from $1000 cm^{-1}$ to $3700 cm^{-1}$, except DFIR which has 220 bands spanning from $1000 cm^{-1}$ to $1760 cm^{-1}$. In this graph, the solid line is the mean of an area in the collagen region of the tissue. The shaded area is the standard deviation of the pixels in the area. The blue graph depicts the DFIR image's spectral response, the grey graph depicts the FTIR image's spectral response, the orange graph depict dictionary learning image's spectral response, the green graph depict LASSO image's spectral response.}
    \label{fig:Collagen1}
\end{figure*}
\subsection{Result Comparison} \label{IV.C}
In this section, we compare the differences in algorithms and fusion metrics between our method, Bayes Sparse and HySure due to their performance results are closer to ours than CSTF and CO-CNMF. Our algorithm outperforms these methods by a significant margin, with performance converging after $10$dB (Table \ref{tab:PaviaTab}). Similarly, our metrics values are much closer to the ideal value (zero) using RMSE, DD, and ERGAS metrics. High performance in the presence of low SNR is critical for expanding FTIR to medical imaging by providing data for a complete biopsy in a couple of hours. While image quality is an important metric, we also evaluate the tensor's pixel-level spectral response using the SAM metric, where our algorithm is able to produce an average of 2.4 for all pixels. The second best algorithm is Bayes Sparse with SAM average value at 4.8. In Table \ref{tab:PaviaTab}, our algorithms has a significantly reduced computational time compared to others at an average of \SI{5.16}{s} in University of Pavia dataset. The second best time belongs to HySure at an average of \SI{38.32}{s}. Algorithm time is measured after HSI and MSI is simulated and fed to algorithms until a fused image is produced. In the case of Bayes Sparse, the Algorithm Time is the slowest of all methods. It takes about \SI{105}{s} to train dictionaries, and about \SI{70}{s} to fuse images.

Figure \ref{fig:Pavia} shows the difference between our results and fusion images produced using competing methods (Bayes Sparse, HySure, CO-CNMF, and CSTF) for a $10$dB SNR and $39\times 39$ Gaussian blur kernel. The two fusion images which are closest to Ground Truth are ours and Bayes Sparse. HySure fusion image has a small amount of artifacts, while CO-CNMF and CSTF fusion image have a significant amount of artifacts. When we ran all algorithms with the same SNR and a $5\times 5$ blur kernel, the artifacts are not present in CO-CNMF, CSTF, and HySure anymore. We concluded that the three algorithms (HySure, CO-CNMF, and CSTF) are not not capable of with resolving HSI's blur effect which HSI suffers seriously from. When we look at the color fidelity in LASSO and Bayes Sparse fusion images, both of them have accurate color reproduction except the area of landmass in the middle of the image, highlighted by Figure \ref{fig:Pavia}'s Best Cases, where ours has slight reddish color and Bayes Sparse has a reddish color instead of bright yellow in Ground Truth.

We suspect HySure's low performance is caused by its inability to adjust weights based on to MSI and HSI noise. We present equation (6) in \cite{simoes2015convex} here:
\begin{multline}
    \min_{\mathbf{R}} \frac{\mu}{2} \left\lVert(\mathbf{H} - \mathbf{LRQ})\right\rVert_F^2 + \frac{\mu}{2}\left\lVert(\mathbf{M} - \mathbf{RQB})\right\rVert_F^2\\
    + \eta\varphi(\mathbf{R}\mathbf{D}_h,\mathbf{R}\mathbf{D}_v).
    \label{eqn:ConvexOp}
\end{multline}
The sparse regularizer is a vector Total Variation represented by the function $\varphi()$. Compared to our minimization functions in Equations \ref{eqn:MaxaP} and \ref{eqn:LagEq}, HySure uses similar scalar multipliers (such as $\mu$ and $\eta$) and a similar optimization scheme. These scalar multipliers act as weights on how important the regularization on those terms as well as defining the threshold for nuclear norm. However, HySure lacks penalty weights inside the Frobenius norms for HS and MS terms, which means the data from HSI has the same contribution as data from MSI leading to a worse fused image. By utilizing these logarithmic scale weights of HSI and MSI, we dictate which image has higher contribution to our fusion image at a spectral range of our choice. 
In dictionary learning method, although we have similar initialization scheme for $\mathbf{R}$ and regularization for Frobenius norm, the sparse regularizer in Bayes Sparse is learned from the Bayesian MAP $\mathbf{\bar{R}}$. A support will be built based on this and it will identify nonzero elements on each row of $\mathbf{R}$, meaning finding nonzero values of a pixel in spectral dimension. Thus, only nonzero elements defined by dictionaries learnt from $\mathbf{\bar{R}}$ will be updated. Bayes Sparse method works well if the correct support is found, but if the support is incorrect it will filter out both signal and noise or leaving noise behind. Instead of assuming $\mathbf{\bar{R}}$ has the correct support, we adopt LASSO's nuclear norm as our sparse regularizer/noise reduction mechanism. In our algorithm, LASSO's Frobenius norms will grasp all the information available from HSI and MSI, then nuclear norm will reduce noise or enforce sparsity. By doing so, we lose less information compared to Bayes Sparse and giving us a globally optimal solution. 

The spectra fidelity in fused images is best demonstrated in Figure \ref{fig:SAM2}. In this figure, SAM metric values at each pixels are graphed. The range of SAM value here is from $0\degree$ to $57\degree$, with $0\degree$ means no spectral distortion between fused image and ground truth at the particular pixel, while $57\degree$ means the angle between the fused image and ground truth at that pixel's spectra is $57\degree$ out of phase. In short, in this figure the darker the image, the less spectral distortion in fused image. Both LASSO and Bayes Sparse fused image is outperformed by HySure at the building with painted metal sheets (area with a highlighted SAM value), where HySure's SAM is at $12.07\degree$,. However, in the rest of the image, our fusion methods outperforms the other two in every  areas: asphalt, meadows, gravel, trees, baresoil, bitumen, self-locking bricks, shadows. Especially in meadows, bitumen, asphalt, self-locking bricks, and trees areas, where our fused image has a mean of SAM at  $1\degree$ and a standard deviation of $0.13\degree$. 

In Figure \ref{fig:PaviaUniMetrics}, we plot the performance in multiple metrics of the fusion methods at each individual band. The behavior of Bayes Sparse is very interesting when looking at these graphs. At earlier bands between 1 and 68, Bayes Sparse performs very well and on par with our method. Bayes Sparse and our methods produce significantly better result than HySure, with PSNR level difference ranging from $3$dB to $10$dB and occasional equalization. In ERGAS, DD and RMSE metric, there is a much wider difference between those two methods and HySure. In UIQI metrics, the differences between HySure, Bayes Sparse, and our proposed approach are reduced. However, Bayes Sparse's performance in all metrics after band 68th are unsteady. This behavior persists in all of the data sets contaminated with Gaussian noise we tested here. We suspect this is due to the dictionaries learned from MS images at bands which lies at the beginning of the spectra. Hence, the data it represents no longer valid at the end of the spectra.

From here on, HySure, CO-CNMF, and CSTF algorithms will not be included because their performance are lacking compared to the our proposed method and Bayes Sparse. In Figure \ref{fig:Tissue2900}, we are showing a set of images from a breast tissue at two different bands. On the first row, images from DFIR imaging system(MSI), FTIR imaging system (HSI), our method (LASSO), Wei's method (Bayes Sparse) are at wavelength \SI{1550}{\centi\meter^{-1}}. Inside the tissue core, there are regions of epithelium and collagen. In this tissue data set, the DFIR's image, Figure \ref{fig:Tissue2900}.a, which shows a good contrast between collagen and epithelium, as well as showing patterns in the collagen and epithelium regions. DFIR's image, MSI, has a spatial resolution of $1128 \times 1152$ pixels. The HSI, Figure \ref{fig:Tissue2900}.b, has a very low spatial resolution at $282 \times 288$ pixels. In HSI, the distinction between collagen and epithelium is clear, but all spatial patterns within the two regions are indistinguishable. The next image in row one is our fusion method's result, Figure \ref{fig:Tissue2900}.c. The contrast in our fusion image is closer to the original image than Bayes Sparse's fusion image. Our fused image has similar patterns to DFIR image in both collagen and epithelium regions. The fusion image from Bayes Sparse method in \cite{wei2015hyperspectral} is on the upper right corner, Figure \ref{fig:Tissue2900}.d. In this image, the contrast is not close to the MSI. Both fusion images have fringing effect, but ours is less than Bayes Sparse's. On the second row, Bayes Sparse fusion image has their contrast distorted compared to HSI. Fringing effect in Bayes Sparse image is significantly worse here while ours is lessen. We suspect again this is due to the dictionaries learned by Bayes Sparse method are from earlier bands in the spectra, which is heavily based on DFIR image which has high spatial resolution with fringing effect. 
 
In Figure \ref{fig:Epithelium1} and \ref{fig:Collagen1}, we compare the mean and standard deviation of spectra from the original and fusion images using our algorithm and Bayes Sparse. Reconstructed spectra represent the mean values of two tissue classes: epithelium and collagen. The performance of our algorithm is similar to Bayes Sparse in the fingerprint region (\numrange{1000}{1700}\si{\centi\meter^{-1}}) where data is available for both FTIR and DFIR source images (HSI and MSI). However, the statistical distribution of spectra in at larger wavenumbers conforms much better to the source FTIR image using the proposed algorithm. Bayes Sparse's statistical deviation corresponds to the error found in the Pavia data set and also correlates with large SAM values. 

Due to the fact that we do not have a ground truth with high resolution in both spatial and spectral domain for this data set, we have to compare the fused image to MSI in \SI{1000}{\centi\meter^{-1}} to \SI{1700}{\centi\meter^{-1}} region, and HSI in the entire spectrum. Table \ref{tab:TissuePerf2} and \ref{tab:TissuePerf1} help us showcase our performance numerically. In Table \ref{tab:TissuePerf2}, the matched wavelengths from \SI{1000}{\centi\meter^{-1}} to \SI{1700}{\centi\meter^{-1}} in HSI, LASSO, and Bayes Sparse images are extracted and compared. In this region, both algorithms perform very well with ours leads on all metrics but UIQI. We leads in the following metrics: PSNR (\SI{3.8}{dB} difference), RMSE ($5\cdot 10^{-3}$ difference), DD ($2.5\cdot 10^{-3}$ difference). This confirms the behaviors of our graphs in Figure \ref{fig:Epithelium1}, where both fused images' spectra follows DFIR's closely. In Table \ref{tab:TissuePerf1}, the fused images are down-sampled by averaging and compared to FTIR image on the entire spectral range. Our algorithm leads by a wide margin here. We leads in the following metrics: PSNR (\SI{4.4}{dB} difference), RMSE ($4.5\cdot 10^{-2}$ difference), UIQI ($0.83\cdot 10^{-3}$ difference), DD ($2.4\cdot 10^{-3}$ difference). The important metrics in this comparison is Degree of Distortion, because this metric's purpose is to evaluate fused image with no ground truth. In both Tables \ref{tab:TissuePerf2} and \ref{tab:TissuePerf1}, the metrics SAM and ERGAS are not included. The tissue image has background regions (dark blue areas in Figure \ref{fig:Tissue2900}) in which pixels values are zero which lead to SAM values at those pixel cannot be calculated. Also, in the spectral range of \SI{1800}{\centi\meter^{-1}} to \SI{2800}{\centi\meter^{-1}}, there are no spectral information so mean values at these wavelengths are close to zero. Therefore, ERGAS values in this region approaches infinity. 

The last thing we want to discuss in this data set's experiment is computational time. The fusion time measures the time to solve the fusion problem only, while algorithm time measures time in all calculations steps after data (HSI and MSI) is fed into algorithms. In Fusion Time, our method requires \SI{354.2}{s} to complete and Bayes Sparse requires \SI{2170.9}{s} to complete. In Algorithm (Alg.) time, our method requires \SI{479.3}{s} to complete and Bayes Sparse requires \SI{6976.9}{s} to complete.
 
\begin{figure*}
  \includegraphics[width=\textwidth]{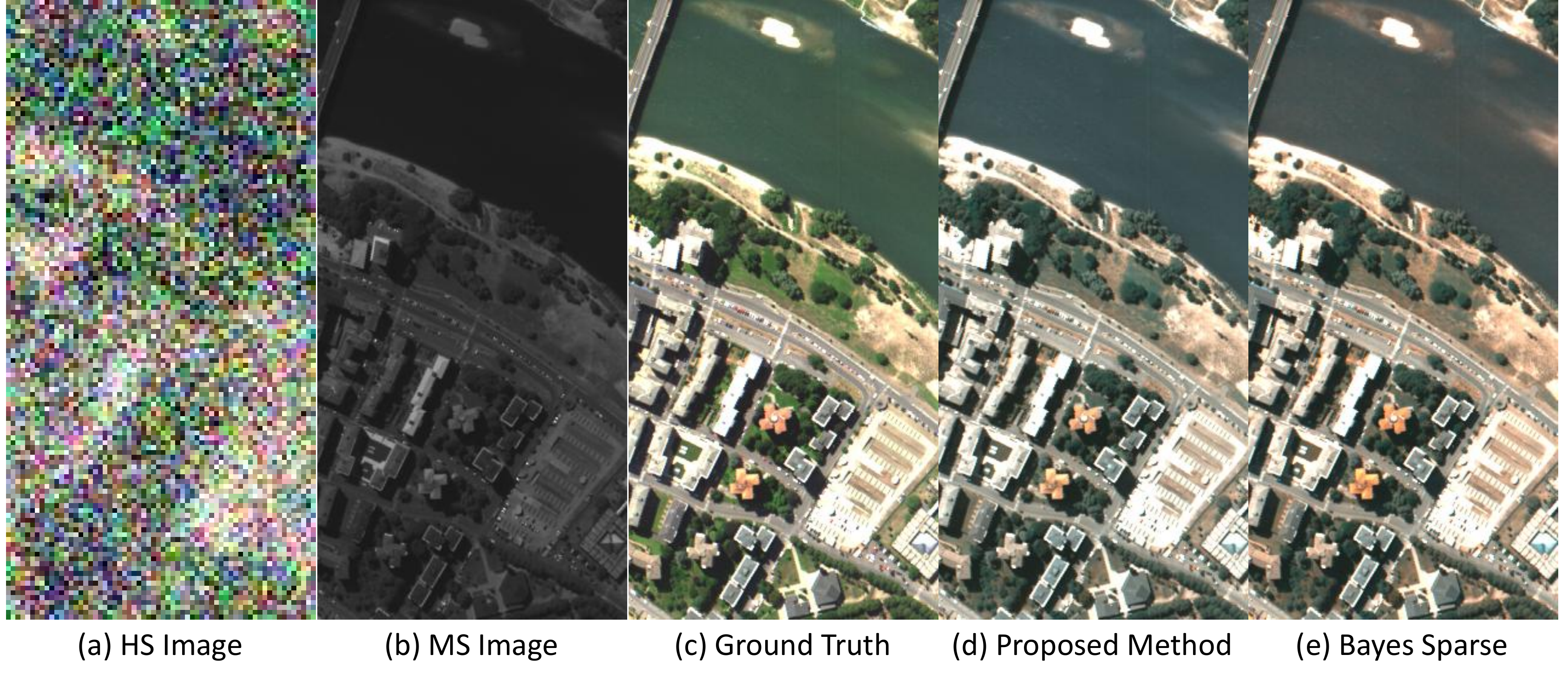}
      \caption{Pavia Center dataset. Blurry HSI was blurred with a 39x39 Gaussian kernel and then down-sampled by every 4 pixels, Poisson noise is added at 10dB SNR (HSI) and 50 dB SNR (MSI).}
  \label{fig:PaviaCenter}
\end{figure*}

\begin{figure*}
  \includegraphics[width=\textwidth]{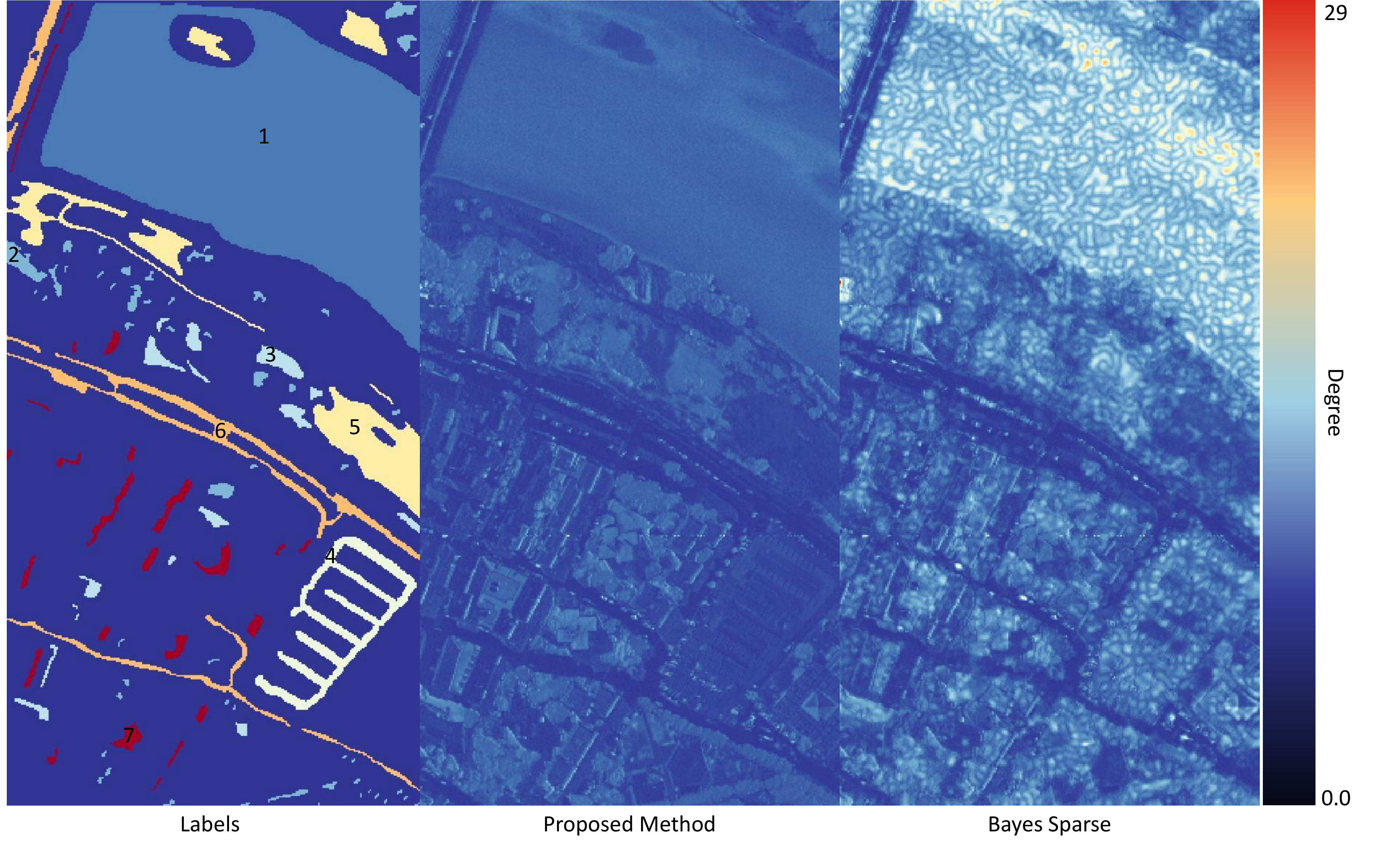}
      \caption{Pavia Center's ground truth labels and spectral angle mapping (SAM) results per pixel, with HSI contaminated with Poisson noise. (a) The ground truth labels provided by expert annotation are shown for context. (b-c) SAM angle at each pixel as a result of image fusion using (b) Proposed method and (c) Bayes Sparse. Ground truth materials are annotated as: (1) water, (2) trees, (3) asphalt, (4) Self-locking bricks, (5) Bitumen, (6) Tiles, (7) Bare soil.}
  \label{fig:SAMPC}
\end{figure*}

\begin{figure}
\centering
  \includegraphics[scale = 0.23]{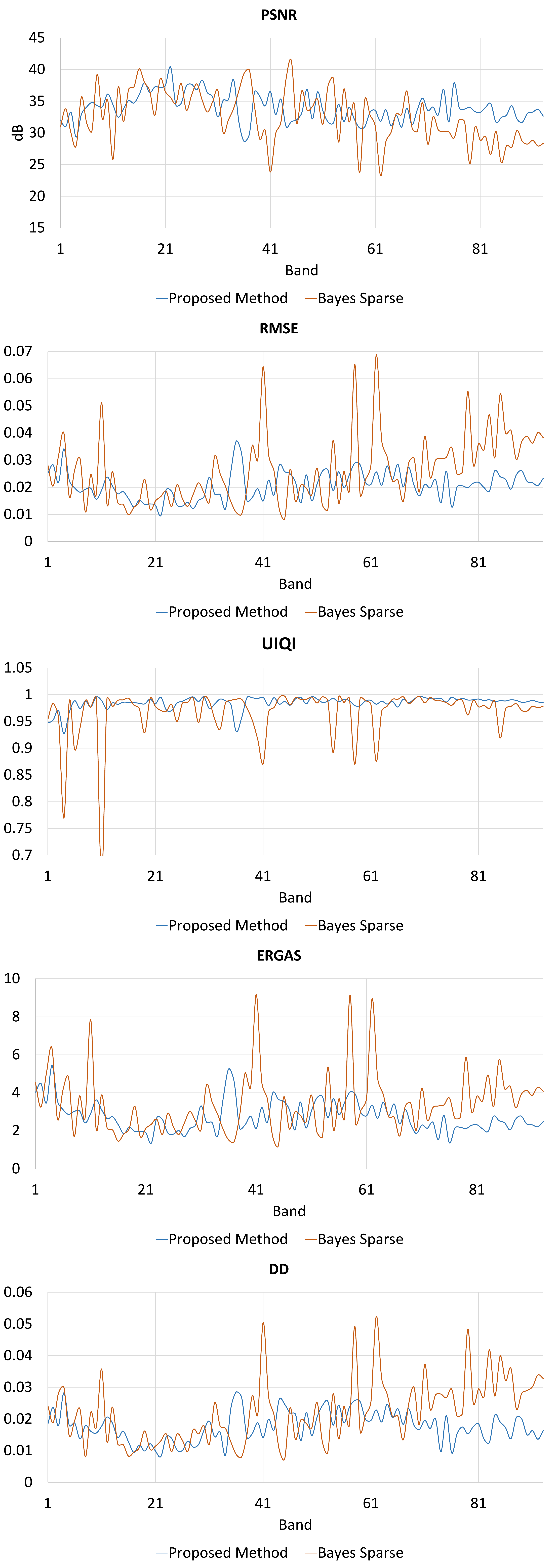}
      \caption{Metrics performance of fusion methods as a function of spectral band (Pavia Center data set with Poisson noise). Top to bottom: PSNR, RMSE, UIQI, ERGAS, DD.}
  \label{fig:PaviaCenterMetricsPoisson}
\end{figure}

\begin{figure}
    \centering
      \includegraphics[scale = 0.23]{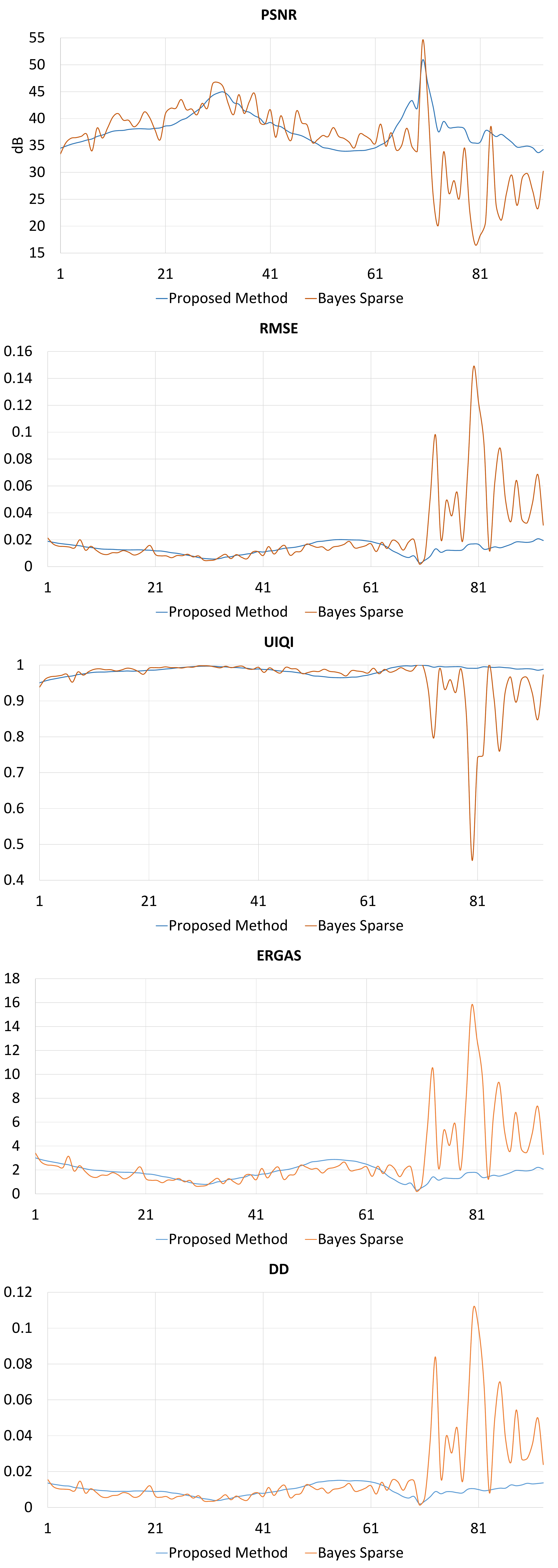}
          \caption{Metrics performance of fusion methods as a function of spectral band (Pavia Center data set with Gaussian noise). Top to bottom: PSNR, RMSE, UIQI, ERGAS, DD.}
      \label{fig:PaviaCenterMetricsGaussian}
\end{figure}

In Figure \ref{fig:PaviaCenter} (Supplemental Materials), we represent Pavia Center data set's fused images when the HSI and MSI are contaminated with Poisson noise at \SI{10}{dB} SNR and \SI{50}{dB} SNR, respectively. Noises were removed from both fused images, but the color fidelity in them is not as good as in our previous Pavia University experiment. In Figure \ref{fig:SAMPC} (Supplemental Materials), the different in spectral quality at each pixel is shown. In all materials, our fused image has much darker color than Bayes Sparse's fused image with an overall average SAM value at $3.25\degree$, while Bayes Sparse's fuse image has an overall average SAM value at $6.46\degree$. In Table \ref{tab:PCPoiTab} (Supplemental Materials), we show our method and Bayes Sprase's performance in different metric for Pavia Center data set with Poisson noise. At all HSI noise level, our proposed method outperforms Bayes Sparse similarly to Table \ref{tab:PaviaTab}, but there is a reduction in fused images' quality in both methods when SNR decreases in HSI. The noise distribution type is the cause of the drop in fusion images' quality. In order to prove this, we ran the experiment with Pavia Center again with Gaussian noise. In Table \ref{tab:PCgauTab} (Supplemental Materials), our method's performance is again consistent at all SNR level in HSI, which is similar to our method's performance in Pavia University data set with Gaussian noise. In PSNR metric, when switching from Gaussian noise to Poisson noise, our method and Bayes Sparse lose an average of \SI{6.66}{dB} and \SI{1.86}{dB}, respectively. In Figure \ref{fig:PaviaCenterMetricsPoisson} (Supplemental Materials), we once again look at the two methods' performance at each band with Pavia Center data set. The noise distribution here is Poisson and it provides us an interesting change in behavior of Bayes Sparse's performance. Bayes Sparse's graphs fluctuates heavily through the entire spectral bands, not just in the later bands. In Figure \ref{fig:PaviaCenterMetricsGaussian} (Supplemental Materials), we show the performance of our method and Bayes Sparse in Pavia Center data set with Gaussian noise. Bayes Sparse graphs fluctuates at the later bands again similar to our experiment with Pavia University data set in Figure \ref{fig:PaviaUniMetrics}.

In Table \ref{tab:MultiDataTab} (Supplemental Materials), we performed similar experiments on Indian Pines (IP), Kennedy Space Center (KSC), and Salinas (Sal.) data sets in order to prove our method's superiority. Both Bayes Sparse and our method are fed with the same sets of HSIs and MSIs under two noise distributions. In Indian Pines and Salinas data sets, our method outperforms again with an average of \SI{4}{dB} difference in PSNR metric. In KSC data set, our sparsest data set, the performance differences between the two fusion methods is negligible. Noise environments still have the same effect on both fusion methods. With Gaussian noise we have an average of \SI{6.2}{dB} difference in PSNR between our method and Bayes Sparse. With Poisson noise we have an average of \SI{1.85}{dB} difference in PSNR between our method and Bayes Sparse.
\section{Conclusion} \label{V}
In this paper, we proposed an approach in image fusion with LASSO. With the implementation of ADMM and Bayesian MAP for initialization, our algorithm obtains a significantly better result when our inputs are badly blurred and heavily damaged with noise. Our method outperforms other fusion methods in every performance metric as well as computing time, which is largely due to our robustness of sparse regularization. 
\section*{Acknowledgments} 
This work was funded in part by the Cancer Prevention and Research Institute of Texas (CPRIT) \#RR140013, the National Institutes of Health / National Library of Medicine \#R00 LM011390-02, Agilent Technologies University Relations Grant \#3938, and the University of Houston Core for Advanced Computing and Data Science (CACDS) SeFAC award.

The authors would like to thank Dr. Qi Wei for publishing the codes of \cite{wei2015hyperspectral}, Dr. Chia-Hsiang Lin for publishing the codes of \cite{8107710}, and Miguel Simoes for publishing the codes of \cite{simoes2015convex}.

\small
\bibliographystyle{IEEEtran}
\bibliography{generic}
\newpage
 \pagenumbering{gobble}
\addcontentsline{toc}{section}{\hspace*{-\tocsep}Acknowledgments} 
\begin{table*}[htbp]
  \centering
  \caption{HS + MS FUSION METHODS PERFORMANCES ON PAVIA CENTER DATA SET WITH POISSON NOISE (PSNR in decibels, SAM in degrees, Algorithm Time in seconds, other metrics are unitless)}
    \begin{tabular}{|c|c|c|c|c|c|c|c|c|}
    \toprule
    HS Noise & Methods & PSNR & RMSE ($10^{-2}$) & SAM & UIQI ($10^{-1}$) & ERGAS & DD ($10^{-2}$) & Alg. Time (s) \\
    \midrule
    \multirow{2}[4]{*}{5 dB} & Bayes Sparse & 24.94 & 5.66  & 12.40 & 8.85  & 7.52  & 4.26  & 171.56 \\
\cmidrule{2-9}          & Proposed & \textbf{27.87} & \textbf{4.04} & \textbf{5.83} & \textbf{9.49} & \textbf{5.42} & \textbf{3.46} & \textbf{5.36} \\
    \midrule
    \multirow{2}[4]{*}{8 dB} & Bayes Sparse & 26.39 & 4.79  & 13.39 & 9.43  & 6.07  & 3.31  & 171.22 \\
\cmidrule{2-9}          & Proposed & \textbf{31.64} & \textbf{2.62} & \textbf{3.94} & \textbf{9.79} & \textbf{3.51} & \textbf{2.21} & \textbf{5.44} \\
    \midrule
    \multirow{2}[4]{*}{10 dB} & Bayes Sparse & 28.62 & 3.71  & 6.46  & 9.51  & 4.70  & 2.49  & 163.80 \\
\cmidrule{2-9}          & Proposed & \textbf{33.41} & \textbf{2.14} & \textbf{3.25} & \textbf{9.85} & \textbf{2.86} & \textbf{1.77} & \textbf{5.86} \\
    \bottomrule
    \end{tabular}%
  \label{tab:PCPoiTab}%
\end{table*}%

\begin{table*}[htbp]
  \centering
      \caption{HS + MS FUSION METHODS PERFORMANCES ON PAVIA CENTER DATA SET WITH GAUSSIAN NOISE (PSNR in decibels, SAM in degrees, Algorithm Time in seconds, other metrics are unitless)}
    \begin{tabular}{|c|c|c|c|c|c|c|c|c|}
    \toprule
    HS Noise & Methods & PSNR & RMSE ($10^{-2}$) & SAM & UIQI ($10^{-1}$) & ERGAS & DD ($10^{-2}$) & Alg. Time (s) \\
    \midrule
    \multirow{2}[4]{*}{5 dB} & Bayes Sparse & 23.49 & 6.69  & 16.65 & 9.07  & 7.25  & 3.18  & 166.43 \\
\cmidrule{2-9}          & Proposed & \textbf{37.28} & \textbf{1.37} & \textbf{3.32} & \textbf{9.84} & \textbf{1.88} & \textbf{0.93} & \textbf{5.46} \\
    \midrule
    \multirow{2}[4]{*}{8 dB} & Bayes Sparse & 27.69 & 4.13  & 10.73 & 9.54  & 4.55  & 2.00  & 169.43 \\
\cmidrule{2-9}          & Proposed & \textbf{37.69} & \textbf{1.30} & \textbf{3.19} & \textbf{9.85} & \textbf{1.82} & \textbf{0.90} & \textbf{5.40} \\
    \midrule
    \multirow{2}[4]{*}{10 dB} & Bayes Sparse & 31.94 & 2.53  & 5.19  & 9.82  & 3.45  & 1.87  & 162.27 \\
\cmidrule{2-9}          & Proposed & \textbf{37.88} & \textbf{1.28} & \textbf{3.05} & \textbf{9.86} & \textbf{1.76} & \textbf{0.88} & \textbf{5.28} \\
    \bottomrule
    \end{tabular}%
  \label{tab:PCgauTab}%
\end{table*}%

\begin{table*}[htbp]
  \centering
  \caption{HS + MS FUSION METHODS PERFORMANCES ON INDIAN PINES, SALINAS, AND KSC DATA SET WITH GAUSSIAN AND POISSON NOISE (PSNR in decibels, SAM in degrees, Algorithm Time in seconds, other metrics are unitless)}
    \begin{tabular}{|c|c|l|c|c|c|c|c|c|c|}
    \toprule
    Dataset & Noise & \multicolumn{1}{c|}{Method} & PSNR & RMSE($10^{-2}$) & SAM & UIQI($10^{-1}$) & ERGAS & DD($10^{-2}$) & Alg. Time (s) \\
    \midrule
    \multirow{4}[8]{*}{IP} & \multirow{2}[4]{*}{Gau.} & Bayes Sparse & 28.10 & 3.94  & 5.89  & 4.18  & 2.13  & 2.08  & 80.87 \\
\cmidrule{3-10}          &       & Proposed & \textbf{31.34} & \textbf{2.71} & \textbf{4.15} & \textbf{5.61} & \textbf{1.52} & \textbf{1.56} & \textbf{1.81} \\
\cmidrule{2-10}          & \multirow{2}[4]{*}{Poi.} & Bayes Sparse & 22.41 & 7.58  & 12.17 & 3.96  & 6.07  & 5.91  & 83.49 \\
\cmidrule{3-10}          &       & Proposed & \textbf{24.67} & \textbf{5.84} & \textbf{9.32} & \textbf{5.06} & \textbf{4.92} & \textbf{4.68} & \textbf{1.87} \\
    \midrule
    \multirow{4}[8]{*}{KSC} & \multirow{2}[4]{*}{Gau.} & Bayes Sparse & 34.61 & 1.86  & 49.45 & \textbf{6.91} & 76.13 & 0.22  & 125.63 \\
\cmidrule{3-10}          &       & Proposed & \textbf{34.89} & \textbf{1.80} & \textbf{16.15} & 5.27  & \textbf{74.11} & \textbf{0.10} & \textbf{2.10} \\
\cmidrule{2-10}          & \multirow{2}[4]{*}{Poi.} & Bayes Sparse & 31.27 & 2.73  & 27.28 & \textbf{1.53} & 159.35 & 1.87  & 106.97 \\
\cmidrule{3-10}          &       & Proposed & \textbf{32.47} & \textbf{2.38} & \textbf{13.99} & 0.52  & \textbf{133.92} & \textbf{1.61} & \textbf{3.00} \\
    \midrule
    \multirow{4}[8]{*}{Sal.} & \multirow{2}[4]{*}{Gau.} & Bayes Sparse & 24.66 & 5.85  & 11.33 & 7.72  & 3.77  & 1.69  & 128.60 \\
\cmidrule{3-10}          &       & Proposed & \textbf{33.83} & \textbf{2.03} & \textbf{4.07} & \textbf{8.16} & \textbf{1.74} & \textbf{0.95} & \textbf{3.36} \\
\cmidrule{2-10}          & \multirow{2}[4]{*}{Poi.} & Bayes Sparse & 28.49 & 3.76  & 8.95  & 5.92  & 6.24  & 2.88  & 126.69 \\
\cmidrule{3-10}          &       & Proposed & \textbf{29.99} & \textbf{3.17} & \textbf{7.28} & \textbf{7.11} & \textbf{5.53} & \textbf{2.48} & \textbf{3.32} \\
    \bottomrule
    \end{tabular}%
  \label{tab:MultiDataTab}%
\end{table*}%
\end{document}